\documentclass[10pt,journal,twocolumn,final,twoside]{IEEEtran}
\usepackage{ifpdf}
\usepackage{cite}
\usepackage{graphicx}
\usepackage[cmex10]{amsmath}
\usepackage{algorithmic}
\usepackage{array}
\usepackage{colortbl}
\usepackage{color}
\usepackage{eqparbox}
\usepackage[tight,footnotesize]{subfigure}
\usepackage{stfloats}
\usepackage{amssymb}
\usepackage{bm}
\usepackage{overpic}
\usepackage{dsfont}
\usepackage{cases}
\usepackage{hyperref}
\usepackage{algorithmic}
\usepackage{algorithm}
\usepackage{multirow}
\usepackage{threeparttable}
\usepackage{cancel}
\usepackage{amsthm}

\newtheorem{assumption}{Assumption}

\newcommand{\vertiii}[1]{{\left\vert\kern-0.25ex\left\vert\kern-0.25ex\left\vert #1 
    \right\vert\kern-0.25ex\right\vert\kern-0.25ex\right\vert}}
    
\def\nn{\nonumber}

\def\Expt{\mathbb{E}}
\def\mb{\mathbb}
\def\defeq{\triangleq}
\def\mc{\mathcal}
\def\col{\mathrm{col}}

\def\one{\mathds{1}}

\def\mN{\mc{N}}
\def\mS{\mc{S}}
\def\mT{\mc{T}}

\def\mW{\mc{W}}

\def\x{\bm{x}}
\def\y{\bm{y}}

\definecolor{BLUE}{rgb}{0,0,1}
\definecolor{orange}{RGB}{255,127,0}
\definecolor{lightblue}{RGB}{140,215,255}

\allowdisplaybreaks

\title{Dictionary Learning over Distributed Models}

\begin{document}

\author{Jianshu~Chen,~\IEEEmembership{Member,~IEEE,}%
	~Zaid~J.~Towfic,~\IEEEmembership{Member,~IEEE,}%
        ~and~Ali~H.~Sayed,~\IEEEmembership{Fellow,~IEEE}
\thanks{
Copyright (c) 2014 IEEE. Personal use of this material is permitted. However, permission to use this material for any other purposes must be obtained from the IEEE by sending a request to pubs-permissions@ieee.org.
}
\thanks{
This work was supported in part by NSF grants CCF-1011918 and ECCS-1407712. A short and preliminary version of this work appears in the conference publication \cite{chen2014icasspdictionary}.}
\thanks{
J. Chen is with Microsoft Research, Redmond, WA, 98052. Email: cjs09@ucla.edu.
}
\thanks{
Zaid J. Towfic is with MIT Lincoln Laboratory, Lexington, MA. Email: ztowfic@ucla.edu.
}
\thanks{
A. H. Sayed is with the Department of Electrical Engineering,
University of California, Los Angeles, CA 90095. Email: sayed@ee.ucla.edu.
This work was completed while J. Chen and Z. J. Towfic were PhD students at UCLA.
}
}


\maketitle

\begin{abstract}

In this paper, we consider learning dictionary models over a network of agents, where each agent is only in charge of a portion of the dictionary elements. This formulation is relevant in Big Data scenarios where large dictionary models may be spread over different spatial locations and it is not feasible to aggregate all dictionaries in one location due to communication and privacy considerations. We first show that the dual function of the inference problem is an aggregation of individual cost functions associated with different agents, which can then be minimized  efficiently by means of diffusion strategies. The collaborative inference step generates dual variables that are used by the agents to update their dictionaries without the need to share these dictionaries or even the coefficient models for the training data.  This is a powerful property that leads to an effective distributed procedure for learning dictionaries over large networks (e.g., hundreds of agents in our experiments). Furthermore, the proposed learning strategy operates in an online manner and is able to respond to streaming data, where each data sample is presented to the network once.
\end{abstract}
\begin{keywords}
Dictionary learning, distributed model, diffusion strategies, dual decomposition, conjugate functions, image denoising, novel document detection, topic modeling, bi-clustering.
\end{keywords}

\section{Introduction and Related Work}
\label{Sec:Intro}

Dictionary learning is a useful procedure by which dependencies among input features can be represented in terms of suitable bases\cite{aharon2006ksvd, chi2013petrels,DictionaryLearningSurvey,elad2006image,mairal2010online,zou2006sparse,shen2008sparse,lee2010biclustering,mairal2008supervised,Kasiviswanathan12}. It has found applications in many machine learning and inference tasks including image denoising\cite{elad2006image,mairal2010online}, dimensionality-reduction \cite{zou2006sparse,shen2008sparse}, bi-clustering \cite{lee2010biclustering}, feature-extraction and classification \cite{mairal2008supervised}, and novel document detection \cite{Kasiviswanathan12}. Dictionary learning usually alternates between two steps: (i) an inference (sparse coding) step and (ii) a dictionary update step. The first step finds a sparse representation for the input data using the existing dictionary by solving, for example, a regularized regression problem, while the second step usually employs a gradient descent iteration to update the dictionary entries.

With the increasing complexity of various learning tasks, it is not uncommon for the size of the learning dictionaries to be demanding in terms of memory and computing requirements. It is therefore important to study scenarios where the dictionary is not necessarily available in a single central location but its components are possibly spread out over multiple locations. This is particularly true in Big Data scenarios where large dictionary components may already be available at separate locations and it is not feasible to aggregate all dictionaries in one location due to communication and privacy considerations. This observation motivates us to examine how to learn a dictionary model that is stored over a network of agents, where each agent is in charge of only a portion of the dictionary elements. Compared with other works, the problem we solve in this article is how to learn a distributed dictionary model, which is, for example, different from the useful work in \cite{Chainais2013learning}  where it is assumed instead that each agent maintains the \emph{entire} dictionary model.

In this paper, we first formulate a general dictionary learning problem, where the residual error function and the regularization function can assume different forms in different applications. As we shall explain, this form turns out not to be directly amenable to distributed implementations. However, when the regularization is strongly convex, we  will show that the problem has a dual function that can be solved in a distributed manner using diffusion strategies \cite{Cattivelli10,sayed2013diffusion,chen2012AllertonLimit,chen2013JSTSPpareto}. In this solution, the agents will not need to share their (private) dictionary elements but only the dual variable. Useful  consensus strategies \cite{kar2008distributed,lee2013distributed,bertsekas1997parallel,tsitsiklis1986distributed} can also be used for the same purpose. However, since it has been shown that diffusion strategies have enhanced stability and learning abilities over consensus strategies \cite{sayed2014proc,sayed2014adaptation,tu2012diffusion}, we will continue our presentation by focusing on diffusion strategies.

We will test our proposed algorithm on two important applications of dictionary learning: (i) novel document detection\cite{Kasiviswanathan12,Aiello2013sensing,Takahashi2014Jan}, and (ii) bi-clustering on microarray data \cite{lee2010biclustering}. A third application related to image denoising is considered in \cite{chen2014icasspdictionary}. In the novel document detection problem \cite{Kasiviswanathan12,Aiello2013sensing,Takahashi2014Jan}, each learner receives documents associated with certain topics, and wishes to determine if an incoming document is associated with a topic that has already been observed in previous data. This application is useful, for example, in finance when a company wishes to mine news streams for factors that may impact stock prices. Another example is the mining of social media streams for topics that may be unfavorable to a company. In these applications, our algorithm is able to perform distributed non-negative matrix factorization tasks, with the residual metric chosen as the Huber loss function \cite{huber1964robust}, and is able to achieve a high area under the receiver operating characteristic (ROC) curve. In the bi-clustering experiment, our algorithm is used to learn relations between genes and types of cancer. From the learned dictionary, the patients are subsequently clustered into groups corresponding to different manifestations of cancer. We show that our algorithm can obtain similar clustering results to those in \cite{lee2010biclustering}, which relies instead on a batched (centralized) implementation.

\begin{table*}[!t]
\renewcommand{\arraystretch}{1.7}
\caption{Examples of tasks solved by the general formulation \eqref{Equ:ProbForm:DictLearn_Objective}--\eqref{Equ:ProbForm:DictLearn_Constraint}. The loss functions $f(u)$ are illustrated in Fig. \ref{Fig:ScalarLossFunctions}. }
\label{Tab:Task}

\centering
\begin{threeparttable}
\begin{tabular}{c||c|c|c|c}
\hline \hline
\rowcolor[gray]{0.9} \rule[-1ex]{0pt}{4ex} \textbf{Tasks} & $f(u)$ & $h_y(y)$ & $h_W(W)$   &  $\mathcal{W}_k$\\ 
\hline
\rule[-1ex]{0pt}{4ex} \textbf{Sparse SVD} & $\frac{1}{2} \|u\|_2^2$ & $\gamma \|y\|_1 + \frac{\delta}{2} \|y\|_2^2$ & 0  & $\left\{W_k: \| [ W_k ]_{:,q} \|_2 \le 1\right\}$  \\ 
\hline 
\rule[-1ex]{0pt}{4ex} \textbf{Bi-Clustering} & $\frac{1}{2}\|u\|_2^2$ & $\gamma \|y\|_1 + \frac{\delta}{2} \|y\|_2^2$ & $\beta \cdot \vertiii{W}_1$ \tnote{a}  & $\left\{W_k: \| [ W_k ]_{:,q} \|_2 \le 1\right\}$  \\ 
\hline 
 \multirow{1}{*}{\textbf{Nonnegative Matrix}} & $\frac{1}{2}\|u\|_2^2$ & $\gamma \|y\|_{1,+} + \frac{\delta}{2} \|y\|_2^2$ \tnote{b} & 0 & $\left\{W_k: \| [ W_k ]_{:,q} \|_2 \le 1, \; W_k \succeq 0\right\}$   \\ \cline{2-5}

\multirow{1}{*}{\textbf{Factorization}} & $\displaystyle\sum_{m=1}^M L(u_m)$ \tnote{c}  & $\gamma \|y\|_{1,+} + \frac{\delta}{2} \|y\|_2^2$ & 0 & $\left\{W_k: \| [ W_k ]_{:,q} \|_2 \le 1, \; W_k \succeq 0\right\}$  \\ 

\hline \hline
\end{tabular} 

\begin{tablenotes}
	\vspace{0.5em}
\item[a] 
The notation $\vertiii{W}_1$ is used to denote the sum of all absolute entries in the matrix $W$: $\vertiii{W}_1 = \sum_{m=1}^M \sum_{q=1}^K |W_{mq}|$, which is different from the conventional matrix $1-$norm defined as the maximum absolute column sum: $\|W\|_1 = \max_{1 \le q \le K} \sum_{m=1}^M |W_{mq}|$.

	\vspace{0.5em}
\item[b]
The notation $\|y\|_{1,+}$ is defined as $\|y\|_{1,+} = \|y\|_1$ if $y \succeq 0$ and $\|y\|_{1,+} = +\infty$ otherwise. It imposes infinite penalty on any negative entry appearing in the vector $y$. Since negative entries are already penalized in $\|y\|_{1,+}$, there is no need to penalize it again in the $\frac{\delta}{2}\|y\|_{2}^2$ term.

\item[c]
		The scalar Huber loss function is defined as $L(u_m) \triangleq \begin{cases}
														\frac{1}{2\eta} u_m^2, & |u_m| < \eta \\
														|u_m| - \frac{\eta}{2}, & \textrm{otherwise}
													 \end{cases}$,
		where $\eta$ is a positive parameter.

\end{tablenotes}

\end{threeparttable}

\end{table*}


The paper is organized as follows. In Section \ref{Sec:ProbForm}, we introduce the dictionary learning problem over distributed models. In Section \ref{Sec:DictLearn}, using the concepts of conjugate function and dual decomposition, we transform the original dictionary learning problem into a form that is amenable to distributed optimization. In Section \ref{Sec:Experiments}, we test our proposed algorithm on two applications. In Section \ref{Sec:Conclusion} we conclude the exposition.



\section{Problem Formulation}
\label{Sec:ProbForm}

\subsection{General Dictionary Learning Problem}

We seek to solve the following general form of a \emph{global} dictionary learning problem over a network of $N$ agents connected by a topology:
	\begin{align}
		\min_{W}	\quad&	\Expt
							\Big[
								f( \x_t - W \y_{t}^o )
								+
								h_y( \y_t^o)
							\Big]
							+
							h_{W}(W)
		\label{Equ:ProbForm:DictLearn_Objective}
							\\
		\mathrm{s.t.}
					\quad&	
							W \in \mW
		\label{Equ:ProbForm:DictLearn_Constraint}
	\end{align}
where $\Expt[\cdot]$ denotes the expectation operator, $\x_t$ is the $M \times 1$ input data vector at time $t$ (we use boldface letters to represent random quantities), $\y_t^o$ is a $K\times 1$ coding vector defined further ahead as the solution to \eqref{Equ:ProbForm:InferenceProblem}, and $W$ is an $M \times K$ dictionary matrix. Moreover, the $q$-th column of $W$, denoted by $[W]_{:,q}$, is called the $q$-th dictionary element (or \emph{atom}), $f(u)$ in \eqref{Equ:ProbForm:DictLearn_Objective} denotes a differentiable convex loss function for the residual error, $h_y(y)$ and $h_W(W)$ are convex (but not necessarily differentiable) regularization terms on $y$ and $W$, respectively,  and $\mW$ denotes the convex constraint set on $W$. Depending on the application problem of interest, there are different choices for $f(u)$, $h_y(y)$, $h_W(W)$ and $\mW$. Table \ref{Tab:Task} lists some typical tasks and the corresponding choices for these functions. In regular dictionary learning \cite{mairal2010online}, the constraint set $\mW$ is
	\begin{align}
		\mW			=		\left\{
								W: \;
								\| [ W ]_{:,q} \|_2 \le 1, \; \forall q
							\right\}
		\label{Equ:ProbForm:W_subUnitNormConstraint}
	\end{align}
and in applications of nonnegative matrix factorization \cite{mairal2010online} and novel document detection (topic modeling) \cite{Kasiviswanathan12}, it is
	\begin{align}
		\mW			=		\left\{
								W: \;
								\| [ W ]_{:,q} \|_2 \le 1, \; W \succeq 0, \; \forall q
							\right\}
		\label{Equ:ProbForm:W_subUnitNormNonnegConstraint}
	\end{align}
where the notation $W \succeq 0$ means each entry of the matrix $W$ is nonnegative. We note that if there is a constraint on $y$, it can be absorbed into the regularization factor $h_y(y)$, by including an indicator function of the constraint into this regularization term. For example, if $y$ is required to satisfy $ y \in \mc{Y} = \{y: 0 \preceq y \preceq \one \}$, where $\one$ denotes the all-one vector, we can modify the original regularization $h_y(y)$ by adding an additional indicator function:
	\begin{align}
		h_y(y)	\leftarrow	h_y(y) + I_{\mc{Y}}(y)
	\end{align}
where the indicator function $I_{\mc{Y}}(y)$ for $\mc{Y}$ is defined as
	\begin{align}
		I_{\mc{Y}}(y)	\defeq		
							\begin{cases}
								0,		&	\mathrm{if~} 0 \preceq y \preceq \one
											\\
								+\infty,	&	\mathrm{otherwise}
							\end{cases}
		\label{Equ:ProbForm:IndicatorFun_def}
	\end{align}
The vector $\y_t^o$ in \eqref{Equ:ProbForm:DictLearn_Objective} is the solution to the following general inference problem for each input data sample $x_t$ at time $t$ for a given $W$ (the regular font for $x_t$ and $y_t^o$ denotes realizations for the random quantities $\x_t$ and $\y_t^o$):
	\begin{align}
		y_t^o		\defeq	\arg\min_{y}
							\left[
								f( x_t - W y ) + h_y(y)
							\right]
		\label{Equ:ProbForm:InferenceProblem}
	\end{align}
Note that dictionary learning consists of two steps: the inference step (sparse coding) for $x_t$ at each time $t$ in \eqref{Equ:ProbForm:InferenceProblem}, and the dictionary update step (learning) in \eqref{Equ:ProbForm:DictLearn_Objective}--\eqref{Equ:ProbForm:DictLearn_Constraint}.

\subsection{Dictionary Learning over Networked Agents}	
\label{Sec:ProbForm:DictLearnNetwork}

%
%
%
%
%
%
%

\begin{figure}
	\centering
	\includegraphics[width=0.3\textwidth]{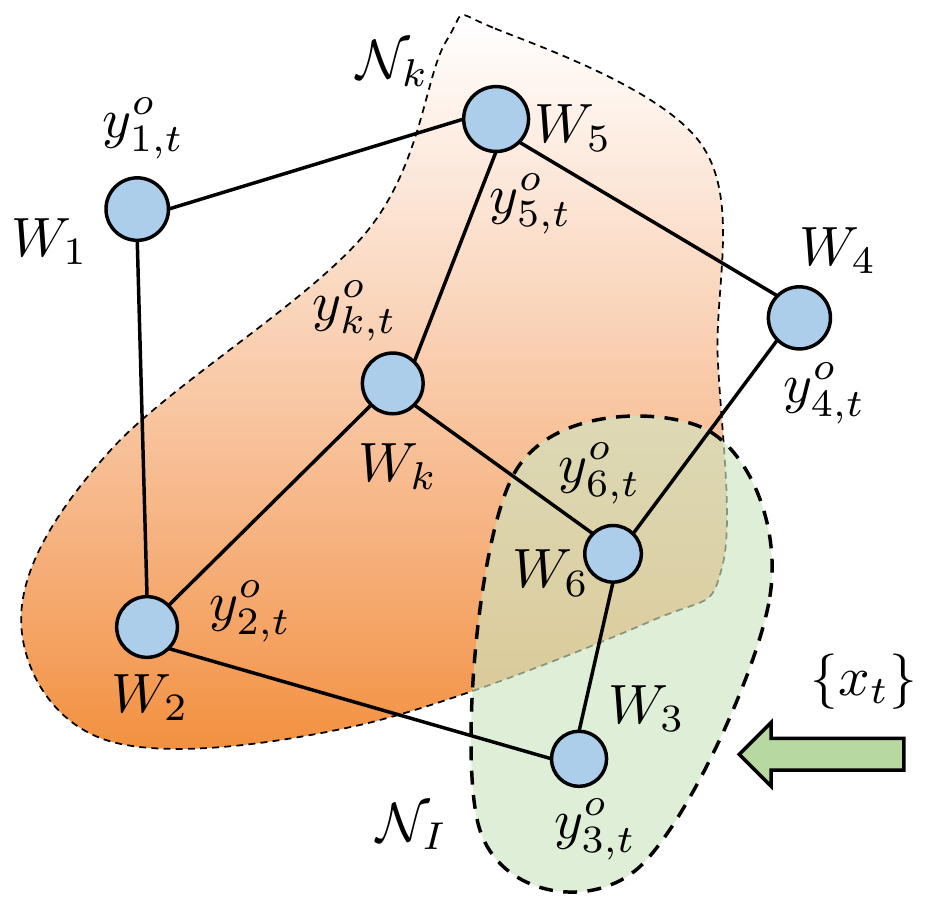}
	\caption{The data sample $x_t$ at time $t$
	is available to	a subset $\mN_I$ of agents in the network (e.g., agents
	$3$ and $6$ in the figure), 
	and each agent $k$ is in charge of one sub-dictionary, $W_k$, and the corresponding optimal 
	sub-vector of coefficients estimated at time $t$, $y_{k,t}^o$. Each agent
	$k$ can only exchange information with its immediate neighbors (e.g.,
	agents $5$, $2$ and $6$ in the figure and $k$ itself). We use $\mN_k$ to denote the 
	set of neighbors of agent $k$. }
	\label{Fig:DistributedModel}
\end{figure}

Let the matrix $W$ and the vector $y$ be partitioned in the following block forms:
	\begin{align}
		W		&=		\begin{bmatrix}
							W_1  & \cdots & W_N
						\end{bmatrix},
						\quad
		y		=		\col\{ y_1, \; \ldots, \; y_N\}
		\label{Equ:ProbForm:W_y_def}
	\end{align}
where $W_k$ is an $M \times N_k$ \emph{sub-dictionary} matrix and $y_k$ is an $N_k \times 1$ sub-vector. 
{ Note that the sizes of the sub-dictionaries add up to the total size of the dictionary, $K$, i.e., 
	\begin{align}
		N_1+\cdots+N_N = K
	\end{align}
}%
Furthermore, we assume the regularization terms $h_y(y)$ and $h_W(W)$ admit the following decompositions:
	\begin{align}
		h_y(y)		&=		\sum_{k=1}^N 
							h_{y_k}(y_k),
							\quad
		h_W(W)		=		\sum_{k=1}^N
							h_{W_k}(W_k)
		\label{Equ:ProbForm:hW_factorization}
	\end{align}
Then, the objective function of the inference step \eqref{Equ:ProbForm:InferenceProblem} can be written as
	\begin{align}
		Q(W, y; x_t)
		\defeq
			f\Big(
				x_t - \sum_{k=1}^N W_k y_k
			\Big)
			+
			\sum_{k=1}^N
			h_{y_k}(y_k)
		\label{Equ:ProbForm:InferenceProblem_FactoredForm}
	\end{align}
We observe from \eqref{Equ:ProbForm:InferenceProblem_FactoredForm} that the sub-dictionary matrices $\{W_k\}$ are linearly combined to represent the input data $x_t$. By minimizing $Q(W,y;x_t)$  over $y$, the first term in \eqref{Equ:ProbForm:InferenceProblem_FactoredForm} helps ensure that the representation error for $x_t$ is small. The second term in \eqref{Equ:ProbForm:InferenceProblem_FactoredForm}, which usually involves a combination of $\ell_1$ and $\ell_2$ measures, as indicated in Table~\ref{Tab:Task}, helps ensure that each of the resulting combination coefficients $\{y_k\}$ is sparse and small. We will make the following assumption regarding $h_{y_k}(y_k)$ throughout the paper
	\begin{assumption}[Strongly convex regularization]
		\label{Assumption:StronglyConvexRegularization_hyk}
		The regularization terms $h_{y_k}(y_k)$ are assumed to be strongly convex for $k=1,\ldots,N$. 
		\hfill \qed
	\end{assumption}
\noindent This assumption will allow us to develop a fully distributed strategy that enables the sub-dictionaries $\{W_k\}$ and the corresponding coefficients $\{y_k\}$ to be stored and learned in a distributed manner over the network; each agent $k$ will infer its own $y_k$ and update its own sub-dictionary $W_k$ with limited interaction with its neighboring agents. Requiring $\{h_{y_k}(y_k)\}$ to be strongly convex is not restrictive since we can always add a small $\ell_2$ regularization term to make it strongly convex. For example, in Table \ref{Tab:Task}, we add an $\ell_2$ term to $\ell_1$ regularization so that the resulting $h_{y_k}(y_k)$ ends up amounting to elastic net regularization, in the manner advanced in \cite{zou2006sparse}.

Figure \ref{Fig:DistributedModel} shows  the assumed configuration of the knowledge and data distribution over the network. The  sub-dictionaries $\{W_k\}$ can be interpreted as the ``wisdom'' that is distributed over the network, and which we wish to combine in a distributed manner to form a greater ``intelligence'' for interpreting the data $\x_t$. Observe that we are allowing $\x_t$ to be observed by only a subset, $\mN_I$, of the agents. By having the dictionary distributed over the agents, we would then like to develop a procedure that enables these networked agents to find the \emph{global} solutions to both the inference problem \eqref{Equ:ProbForm:InferenceProblem} and the learning problem \eqref{Equ:ProbForm:DictLearn_Objective}--\eqref{Equ:ProbForm:DictLearn_Constraint} with interactions that are limited to their neighborhoods.

\begin{figure}[t!]
	\centering
	\includegraphics[width=0.48\textwidth]{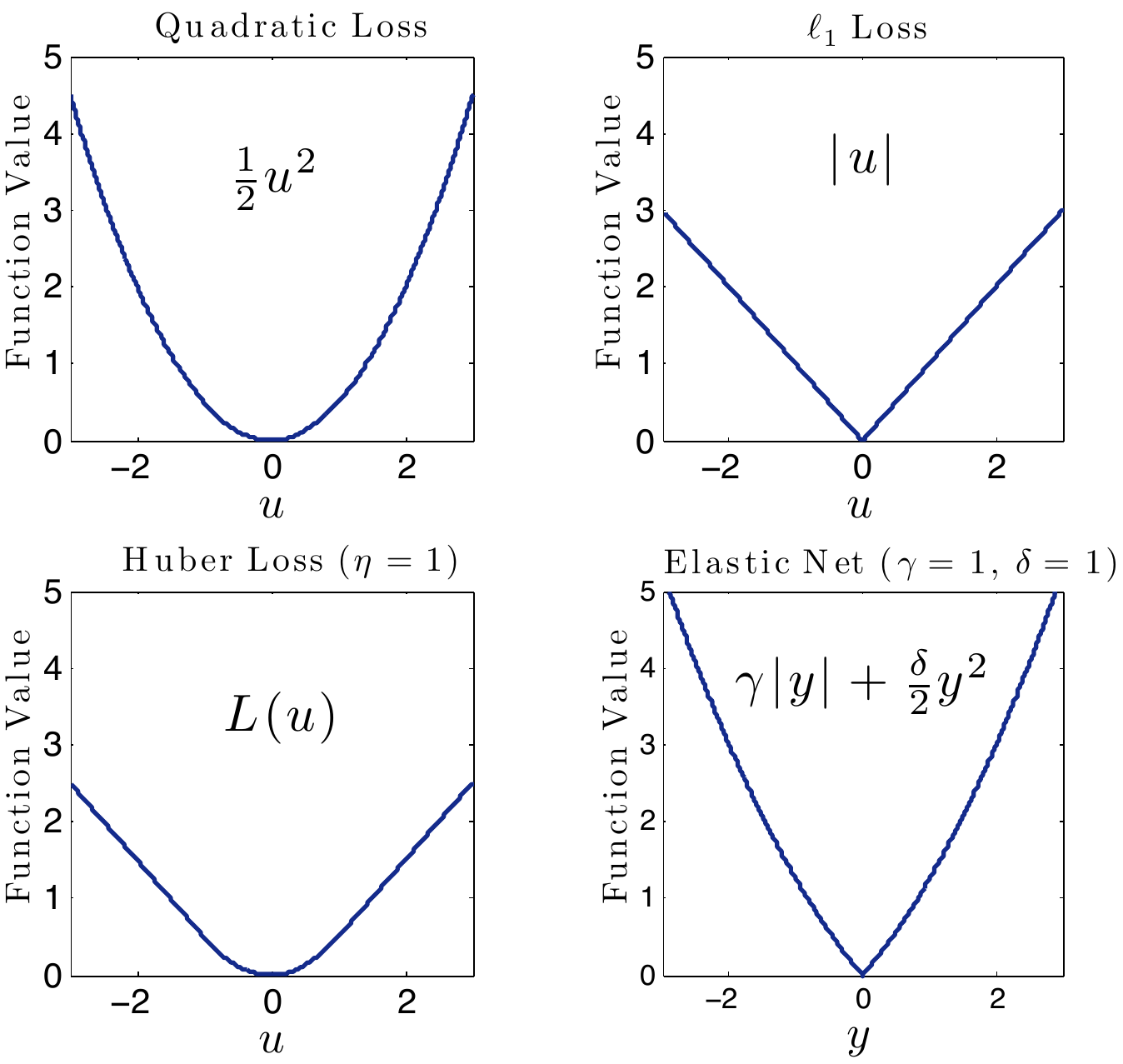}
	\caption{Illustration of the loss functions, and the elastic net regularization.}
	\label{Fig:ScalarLossFunctions}
\end{figure}

\subsection{Relation to Prior Work}
\label{Sec:ProbForm:RelatedWork}

\subsubsection{Model Distributed vs. Data Distributed}
The problem we are solving in this paper is different from the useful work \cite{chainais2013distributed,Chainais2013learning} on distributed dictionary learning and from the traditional distributed learning setting \cite{Cattivelli10,chen2013JSTSPpareto,sayed2013diffusion,chouvardas2011adaptive}, where it is assumed that the \emph{entire} dictionary $W$ is maintained by each agent or that individual data samples generated by the same distribution, denoted by $\x_{k,t}$, are observed by the agents at each time $t$. That is, these previous works study \emph{data distributed} formulations. What we are studying in this paper is to find a distributed solution where each agent is only in charge of a \emph{portion} of the dictionary ($W_k$ for each agent $k$) and where the incoming data, $\x_t$, is observed by only a subset of the agents. This scenario corresponds to a \emph{model distributed} (or dictionary-distributed) formulation. A different formulation is also considered  in \cite{dean2013large} in the context of distributed deep neural network (DNN) models over computer networks. In these models, each computer is in charge of a portion of neurons in the DNN, and the computing nodes exchange their private activation signals. As we will see further ahead, our distributed model requires exchanging neither the private combination coefficients $\{y_k\}$ nor the sub-dictionaries $\{W_k\}$.

The distributed-model setting we are studying is important in practice because agents tend to be limited in their memory and computing power and they may not be able to store large dictionaries locally. Even if the agents were powerful enough, different agents may still have access to different databases and different sources of information. Rather than aggregate the information in the form of large dictionaries at every single location, it is often more advantageous to keep the information distributed due to costs in exchanging large dataset and dictionary models, and also due to privacy considerations where agents may not be in favor of sharing their private information.


\subsubsection{Distributed Basis Pursuit}
\label{Sec:RelatedWork:DBP}
{
Other useful related works appear in the studies \cite{mota2012distributedBP,mota2013DADMM,yuan2013convergence} on distributed basis pursuit, which also rely on dual decomposition arguments. However, there are some key differences in problem formulation, generality, and technique, as explained in \cite{towfic2014dictionary}. For example, the works \cite{mota2012distributedBP,mota2013DADMM,yuan2013convergence}  do not deal with dictionary learning problems and focus instead on the solution of special cases of the inference problem \eqref{Equ:ProbForm:InferenceProblem}. Specifically, the problem formulations in \cite{mota2012distributedBP,mota2013DADMM,yuan2013convergence} focus on determining sparse solutions to (underdetermined) linear systems of equations, which can be interpreted as corresponding to scenarios where the dictionaries are \emph{static}  and not learned from data. In comparison, in this article, we show how the inference {\em and}
{ 
learning problems \eqref{Equ:ProbForm:InferenceProblem} 
and \eqref{Equ:ProbForm:DictLearn_Objective}--\eqref{Equ:ProbForm:DictLearn_Constraint}
}%
can be \emph{jointly} integrated into a common framework. Furthermore, our proposed distributed dictionary learning strategy is an \emph{online} algorithm, which updates the dictionaries sequentially in response to streaming data. We also only require the data sample $x_t$ to be available to a subset of the agents (e.g., one agent) while it is assumed in \cite{mota2012distributedBP,mota2013DADMM,yuan2013convergence} that all agents have access to the same data $x_t$.

For instance, one of the problems studied in \cite{mota2012distributedBP} is the following inference problem (compare with \eqref{Equ:ProbForm:InferenceProblem}):
\begin{subequations}
\label{Equ:ProbForm:motaBP_noisefree}
\begin{align}
		y_t^o		\triangleq	\underset{y}{\arg\min} \quad&
								\sum_{k=1}^N \left[\gamma\|y_k\|_1 + \frac{\delta}{2} \|y_k\|_2^2\right]
		\label{Equ:ProbForm:mota1_cost}
								\\
								\mathrm{s.t.}\quad &\sum_{k=1}^N W_k y_k = x_t
		\label{Equ:ProbForm:mota1}
	\end{align}
\end{subequations}
This formulation can be recast as a special case of \eqref{Equ:ProbForm:InferenceProblem} by selecting:
\begin{subequations}
\begin{align}
	h_{y_k}(y_k) 	&= 	\gamma \|y_k\|_1 + \frac{\delta}{2} \|y_k\|_2^2 
					\label{eq:mota1_regularizer}\\
	f(x_t-Wy) 		&= 	I_{\mc{B}}\Big(x_t-\sum_{k=1}^N W_k y_k\Big) 
					\label{eq:mota1_residual}
\end{align}
\end{subequations}
where $I_{\mc{B}}(\cdot)$ is the indicator function defined by:
\begin{align}
	I_{\mc{B}}(u) = \begin{cases}
							0, & u \in \mc{B}\\
							\infty, & u \notin \mc{B}
						\end{cases} \label{eq:indicator_function}
\end{align}
where $\mc{B} \triangleq \left\{0_M\right\}$ is a set consisting of the zero vector in $\mb{R}^M$. Equality constraints of the form \eqref{Equ:ProbForm:mota1}, or a residual function of the form \eqref{eq:mota1_residual}, are generally problematic for problems that require {\em both} learning and inference since modeling and measurement errors usually seep into the data and the $\{W_k\}$ may not be able to represent the $x_t$ accurately with a precise equality as in \eqref{Equ:ProbForm:mota1}. To handle the modeling error, the work \cite{mota2013DADMM} considered instead:
\begin{subequations}
\label{Equ:ProbForm:motaBP_noisy}
\begin{align}
		y_t^o		\triangleq	\underset{y}{\arg\min} \quad&
								\sum_{k=1}^N \left[\gamma\|y_k\|_1 + \frac{\delta}{2} \|y_k\|_2^2\right] \label{Equ:ProbForm:mota2_lasso_obj}\\
								\mathrm{s.t.}\quad &\Big\|\sum_{k=1}^N W_k y_k - x_t\Big\|_2 \leq \sigma
		\label{Equ:ProbForm:mota2_lasso}
	\end{align}
\end{subequations}
for some $\sigma \geq 0$, which again can be viewed as a special case of problem \eqref{Equ:ProbForm:InferenceProblem} for the same $h_{y_k}(\cdot)$ from \eqref{eq:mota1_regularizer} and with the indicator function in \eqref{eq:mota1_residual} replaced by $I_{\mc{C}}(u)$ relative to the set
\begin{align}
	\mc{C} \triangleq \left\{u \in \mathbb{R}^{M\times 1} : \|u\|_2 \leq \sigma \right\}
	\label{Equ:ProbForm:C_set}
\end{align}
An alternative problem formulation that removes the indicator functions is considered in \cite{mateos2010distributed,mota2013DADMM}, namely, 
\begin{align}
		y_t^o		\triangleq	\underset{y}{\arg\min}
							\left[
								\frac{1}{2} \| x_t - W y \|^2 + \gamma \|y\|_1
							\right]
		\label{Equ:ProbForm:InferenceProblem_mota2_BPDN}
\end{align}
Here, we now have $h_y(y)=\gamma \|y\|_1$ and $f(u)=\frac{1}{2}\|u\|^2$. However, for problem \eqref{Equ:ProbForm:InferenceProblem_mota2_BPDN}, the dictionary elements as well as the entries of $x_t$, were partitioned in \cite{mateos2010distributed,mota2013DADMM} by {\em rows} across the network as opposed to our column-wise partitioning in \eqref{Equ:ProbForm:W_y_def}:
\begin{align}
	W = [U_1^T, \ldots, U_N^T]^T
\end{align}
In this case, it is straightforward to rewrite problem \eqref{Equ:ProbForm:InferenceProblem_mota2_BPDN} in the form 
\begin{align}
	y_t^o		\triangleq	\underset{y}{\arg\min} \sum_{k=1}^N 
							\left[
								\frac{1}{2} \| x_{k,t} - U_k y \|^2 + \frac{\gamma}{N} \|y\|_1
							\right]
\end{align}
which is naturally in a ``sum-of-costs'' form; such forms are directly amenable to distributed optimization and do not require transformations --- see \eqref{Equ:DictLearnDist:J_glob_sumOFcosts} further ahead. However, the more challenging problem where the matrix $W$ is partitioned column-wise as in \eqref{Equ:ProbForm:W_y_def}, which leads to the ``cost-of-sum'' form showed earlier in \eqref{Equ:ProbForm:InferenceProblem_FactoredForm}, was not examined in \cite{mota2013DADMM,mateos2010distributed}. 

In summary, we will solve the more challenging problem of \emph{joint} inference and dictionary learning (instead of inference alone under static dictionaries) under the \emph{column-wise} partitioning of $W$ (rather than row-wise partitioning) and general penalty functions $f(\cdot)$ and $\{h_{y_k}(\cdot)\}$ (instead of the special indicator choices in \eqref{eq:indicator_function} and \eqref{Equ:ProbForm:C_set}).
}

\section{Learning over Distributed Models}
\label{Sec:DictLearn}

\subsection{``Cost-of-Sum'' vs. ``Sum-of-Costs''}
\label{Sec:DictLearn:CostSum}

We thus start by observing that the cost function \eqref{Equ:ProbForm:InferenceProblem_FactoredForm} is a regularized ``\emph{cost-of-sum}''; it consists of two terms: the first term has a sum of quantities associated with different agents appearing as an argument for the function $f(\cdot)$ and the second term is a  collection of separable regularization terms $\{h_{y_k}(y_k)\}$. This formulation is different from the classical ``\emph{sum-of-costs}'' problem, which usually seeks to minimize a global cost function, $J^{\mathrm{glob}}(w)$, that is expressed as the aggregate sum of individual costs $\{J_k(w)\}$, say, as:
	\begin{align}
		J^{\mathrm{glob}}(w)		=	\sum_{k=1}^N J_k(w)
		\label{Equ:DictLearnDist:J_glob_sumOFcosts}
	\end{align}
The ``sum-of-costs'' problem \eqref{Equ:DictLearnDist:J_glob_sumOFcosts} is amenable to distributed implementations\cite{Cattivelli10,sayed2013diffusion,chen2012AllertonLimit,chen2013JSTSPpareto,kar2008distributed,lee2013distributed,bertsekas1997parallel,tsitsiklis1986distributed,sayed2014proc}. In comparison, minimizing the regularized ``cost-of-sum'' problem in \eqref{Equ:ProbForm:InferenceProblem_FactoredForm} directly would require knowledge of all sub-dictionaries $\{W_k\}$ and coefficients $\{y_k\}$. Therefore, this formulation is not directly amenable to the distributed techniques from \cite{Cattivelli10,sayed2013diffusion,chen2012AllertonLimit,chen2013JSTSPpareto,kar2008distributed,lee2013distributed,bertsekas1997parallel,tsitsiklis1986distributed,sayed2014proc}. In \cite{chang2013distributed}, the authors proposed a useful consensus-based primal-dual perturbation method to solve a similar constrained ``cost-of-sum'' problem for smart grid control. In their method, an averaging consensus step was used to compute the sum inside the cost. We follow a different route and arrive at a more efficient distributed strategy by transforming the original optimization problem into a dual problem that has the same form as \eqref{Equ:DictLearnDist:J_glob_sumOFcosts} --- see \eqref{Equ:DictLearnDist:DualProblem_Objective_newForm}--\eqref{Equ:DictLearnDist:DualProblem_Constraint_newForm} further ahead, and which can then be solved efficiently by means of diffusion strategies. There will be no need to exchange any information among the agents beyond the dual variable, or to employ a separate consensus step to evaluate the sum inside the cost in order to update their own sub-dictionaries.

\subsection{Inference over Distributed Models: A Dual Formulation}
\label{Sec:DictLearn:Inference_Dual}

To begin with, we first transform the minimization of \eqref{Equ:ProbForm:InferenceProblem_FactoredForm} into the following equivalent optimization problem { by introducing a splitting variable $z$}:
	\begin{subequations}
		\begin{align}
			\min_{ \{y_k\}, z }		\quad&		f( x_t - z ) + \sum_{k=1}^N h_{y_k}(y_k)
			\label{Equ:DictLearnDist:NewInference_Objective}
												\\
			\mathrm{s.t.}			\quad&		z = \sum_{k=1}^N W_k y_k
			\label{Equ:DictLearnDist:NewInference_Constraint}
		\end{align}
	\end{subequations}
Note that the above problem is convex over both $\{y_k\}$ and $z$ since the objective is convex and the equality constraint is linear. Problem \eqref{Equ:DictLearnDist:NewInference_Objective}--\eqref{Equ:DictLearnDist:NewInference_Constraint} is a convex optimization problem with linear constraints so that strong duality holds\cite[p.514]{bertsekas1999nonlinear}, meaning that the optimal solution to \eqref{Equ:DictLearnDist:NewInference_Objective}--\eqref{Equ:DictLearnDist:NewInference_Constraint} can be found by solving its corresponding dual problem (see \eqref{Equ:DictLearnDist:DualProblem} below) and then recovering the optimal primal variables $\{y_k\}$ and $z$ (to be discussed in Sec. \ref{Sec:DictLearn:RecoveryPrimalVar}):
	\begin{align}
		\max_{\nu} g(\nu; x_t)
		\label{Equ:DictLearnDist:DualProblem}
	\end{align}
where $g(\nu; x_t)$ is the dual function associated with the optimization problem \eqref{Equ:DictLearnDist:NewInference_Objective}--\eqref{Equ:DictLearnDist:NewInference_Constraint}, and is defined as follows. First, the Lagrangian $L(\{y_k\}, z, \nu; x_t)$  over the primal variables  $\{y_k\}$ and $z$ is given by
	\begin{align}
		L&( \{y_k\}, z, \nu; x_t )
				\nn\\
			&=	
				f( x_t - z ) + \nu^T z
				+
				\sum_{k=1}^N
				\Big[
					h_{y_k}(y_k) 
					-
					\nu^T W_k y_k
				\Big]
		\label{Equ:DictLearnDist:Lagrangian}
	\end{align}
Then, the dual function $g(\nu; x_t)$ can be expressed as:
	\begin{align}
		g&(\nu;x_t) \nn\\		&\defeq		
									\inf_{\{y_k\}, z}
									L(\{y_k\}, z, \nu; x_t)
									\nn\\
						&=	
									\inf_{ z } \!
									\left[
										f( x_t \!-\! z ) \!+\! \nu^T z
									\right]
									\!+\!\!
									\sum_{k=1}^N
									\inf_{y_k} \!
										\Big[
											h_{y_k}(y_k) 
											\!-\!
											\nu^T W_k y_k
										\Big]
		\label{Equ:DictLearnDist:DualFunction_def}
									\\
						&\overset{(a)}{=}
								\inf_{u} 
								\left[
									f(u) \!-\! \nu^T u \!+\! \nu^T x_t
								\right]
								\!+\!
								\sum_{k=1}^N
											\inf_{y_k} \!
												\Big[
													h_{y_k}(y_k) 
													\!-\!
													\nu^T W_k y_k
												\Big]
								\nn\\
						&=	
								-\!
								\sup_{u}
								\left[
									\nu^T u \!-\! f(u)
								\right]
								\!+\!
								\nu^T x_t
								\!-\!\!
								\sum_{k=1}^N \!
									\sup_{y_k}
									\left[
										\nu^T W_k y_k \!-\! h_{y_k}\!(y_k)
									\right]
								\nn\\
						&=	
								- f^{\star}(\nu) + \nu^T x_t
								-
								\sum_{k=1}^N h_{y_k}^{\star}(W_k^T \nu)
		\label{Equ:DictLearnDist:DualFunction_expr}
								\\
								&
								\qquad\qquad
								\nu \in \mc{V}_f \cap \mc{V}_{h_{y_1}} \cap \cdots \cap \mc{V}_{h_{y_N}} 
								\nn
	\end{align}
where in step (a) we introduced $u \defeq x_t - z$,  and $f^{\star}(\cdot)$ and $h_{y_k}^{\star}(\cdot)$ are the conjugate functions of $f(\cdot)$ and $h_{y_k}(\cdot)$, respectively, with the corresponding domains denoted by $\mc{V}_f$ and $\mc{V}_{h_{y_k}}$, respectively. We note that the conjugate function { (or \emph{Legendre-Fenchel} transform\cite[p.37]{urruty1993convex2})}, $r^{\star}(\nu)$, for a function $r(x)$  is defined as \cite[pp.90-95]{boyd2004convex}:
	\begin{align}
		r^{\star}(\nu)		\defeq		\sup_{x} 
										\left[
											\nu^T x - r(x)
										\right]
										,
										\quad
										\nu \in \mc{V}_r
		\label{Equ:DictLearnDist:r_conj_def}
	\end{align}
where the domain $\mc{V}_r$ is defined as the set of $\nu$ where the above supremum is finite. { The conjugate function $r^{\star}(\nu)$ and its domain $\mc{V}_r$ are convex regardless of whether $r(x)$ is convex or not \cite[p.530]{bertsekas1999nonlinear}\cite[p.91]{boyd2004convex}.} In particular, it holds that $\mc{V}_r = \mb{R}^M$ if $r(x)$ is strongly convex \cite[p.82]{urruty1993convex2}. Now since $h_{y_k}(\cdot)$ is assumed in Assumption \ref{Assumption:StronglyConvexRegularization_hyk} to be strongly convex, its domain $\mc{V}_{h_{y_k}}$ is the entire $\mb{R}^M$. If $f(u)$ happens to be strongly convex (rather than only convex, e.g., if $f(u)=\frac{1}{2}\|u\|_2^2$), then $\mc{V}_{f}$ would also be $\mb{R}^M$, otherwise it is a convex subset of $\mb{R}^M$. Therefore, the dual function in \eqref{Equ:DictLearnDist:DualFunction_expr} becomes
	\begin{align}
		g(\nu; x_t)		&=		- f^{\star}(\nu) + \nu^T x_t 
								- 
								\sum_{k=1}^N h_{y_k}^{\star}(W_k^T \nu), 
								\; \nu \in \mc{V}_{f}
		\label{Equ:DictLearn:g_expression_final}
	\end{align}
Now, maximizing $g(\nu; x_t)$ is equivalent to minimizing $-g(\nu; x_t)$ so that the dual problem \eqref{Equ:DictLearnDist:DualProblem} is equivalent to
	\begin{subequations}
	\label{Equ:DictLearn:Inference_Dual}
		\begin{align}
			\min_{\nu} 		\quad&		-g(\nu; x_t)
										=
										f^{\star}(\nu) - \nu^T x_t 
										+
										\sum_{k=1}^N h_{y_k}^{\star}(W_k^T \nu)
			\label{Equ:DictLearn:Inference_Dual_Objective}
										\\
			\mathrm{s.t.}	\quad&		\nu 
										\in 
										\mc{V}_{f}
			\label{Equ:DictLearn:Inference_Dual_Constraint}
		\end{align}
	\end{subequations}
Note that the objective function in the above optimization problem is an aggregation of (i) individual costs associated with sub-dictionaries at different agents (last term in \eqref{Equ:DictLearn:Inference_Dual_Objective}), (ii) a term associated with the data sample $x_t$ (second term in \eqref{Equ:DictLearn:Inference_Dual_Objective}), and (iii) a term that is the conjugate function of the residual cost (first term in \eqref{Equ:DictLearn:Inference_Dual_Objective}). In contrast to \eqref{Equ:ProbForm:InferenceProblem_FactoredForm}, the cost function in \eqref{Equ:DictLearn:Inference_Dual_Objective} is now in a form that is amenable to distributed processing. In particular, diffusion strategies \cite{chen2011TSPdiffopt,sayed2013diffusion,sayed2014proc}, consensus strategies \cite{kar2008distributed,lee2013distributed,bertsekas1997parallel,tsitsiklis1986distributed}, or ADMM strategies \cite{mota2012distributedBP,mota2013DADMM,schizas2008consensus1,zhu2009distributed,ling2012multi,towfic2014dictionary} can now be applied 
to obtain the optimal dual variable $\nu_t^o$ in a distributed manner at the various agents. 

To arrive at the distributed solution, we proceed as follows. We denote the set of agents that observe the data sample $x_t$ by $\mN_I$. Motivated by \eqref{Equ:DictLearn:Inference_Dual_Objective}, with each agent $k$, we associate the local cost function:
	\begin{align}
		J_k(\nu; x_t)		
					&\defeq	\!
							\begin{cases}
								- \frac{ \nu^T x_t }{|\mN_I|}
								\!+\! 
								\frac{1}{N} f^{\star}(\nu) 
								\!+\! 
								h_{y_k}^{\star}(W_k^T \nu),
											&		k \in \mN_I		\\
								\frac{1}{N} f^{\star}(\nu) 
								\!+\! 
								h_{y_k}^{\star}(W_k^T \nu),
											&		k \notin \mN_I	
							\end{cases} \label{Equ:DictLearn:Split_Nodes} \!\!
	\end{align}
where $| \mN_I |$ denotes the cardinality of $\mN_I$. Then, the optimization problem \eqref{Equ:DictLearn:Inference_Dual_Objective}--\eqref{Equ:DictLearn:Inference_Dual_Constraint} can be rewritten as
	\begin{subequations}
		\begin{align}
			\min_{\nu}		\quad&		\sum_{k=1}^N J_k(\nu; x_t)		
			\label{Equ:DictLearnDist:DualProblem_Objective_newForm}
									\\
			\mathrm{s.t.}	\quad&		\nu \in \mc{V}_{f}
			\label{Equ:DictLearnDist:DualProblem_Constraint_newForm}
		\end{align}
	\end{subequations}
In Sections \ref{Sec:DictLearn:Inference_Diffusion} and \ref{Sec:DictLearn:Inference_ADMM}, we will first discuss the solution of \eqref{Equ:DictLearnDist:DualProblem_Objective_newForm}--\eqref{Equ:DictLearnDist:DualProblem_Constraint_newForm} for the optimal dual variable, $\nu_t^o$, in a distributed manner. And then in Sec. \ref{Sec:DictLearn:RecoveryPrimalVar}, we will reveal how to recover the optimal primal variables $y_{k,t}^o$ and $z_t^o$ from $\nu_t^o$.

\subsection{Inference over Distributed Models: Diffusion Strategies}
\label{Sec:DictLearn:Inference_Diffusion}

Note that the new equivalent form \eqref{Equ:DictLearnDist:DualProblem_Objective_newForm} is an aggregation of individual costs associated with different agents; each cost $J_k(\nu; x_t)$ only requires knowledge of $W_k$. Consider first the case in which $f(u)$ is strongly convex. Then, it holds that $\mc{V}_f = \mb{R}^M$ and problem \eqref{Equ:DictLearnDist:DualProblem_Objective_newForm}--\eqref{Equ:DictLearnDist:DualProblem_Constraint_newForm} becomes an unconstrained optimization problem of the same general form as problems studied in \cite{chen2013JSTSPpareto,chen2012AllertonLimit}.  Therefore, we can directly apply the diffusion strategies developed in these works to solve \eqref{Equ:DictLearnDist:DualProblem_Objective_newForm}--\eqref{Equ:DictLearnDist:DualProblem_Constraint_newForm} in a fully distributed manner. The adapt-then-combine (ATC) implementation  of the diffusion algorithm then takes the following form:
	\begin{subequations}
		\begin{align}
			\psi_{k,i}		&=		\nu_{k,i-1} - \mu \cdot \nabla_{\nu} J_k( \nu_{k,i-1}; x_t )
			\label{Equ:DictLearnDist:ATC_adapt}
									\\
			\nu_{k,i}		&=		\sum_{ \ell \in \mN_k } a_{\ell k}  \psi_{\ell,i}
			\label{Equ:DictLearnDist:ATC_combine}
		\end{align}
	\end{subequations}
where $\nu_{k,i}$ denotes the estimate of the optimal $\nu_t^o$ at agent $k$ at iteration $i$ (we will use $i$ to denote the $i$-th iteration of the inference, and use $t$ to denote the $t$-th data sample), $\psi_{k,i}$ is an intermediate variable, $\mN_k$ denotes the neighborhood of agent $k$, $\mu$ is the step-size parameter chosen to be a small positive number, and $a_{\ell k}$ is the combination coefficient that agent $k$ assigns to the information received from agent $\ell$ and it satisfies
	\begin{align}
		\sum_{\ell \in \mN_k} a_{\ell k} = 1, 
		\;\; 
		a_{\ell k} > 0 \mathrm{~if~} \ell \in \mN_k,\;\;
		a_{\ell k} = 0 \mathrm{~if~} \ell \notin \mN_k	
		\label{Equ:DictLearnDist:A_condition_general}			
	\end{align}
Let $A$ denote the $N \times N$ matrix that collects $a_{\ell k}$ as its $(\ell,k)$-th entry. 
Then, it is shown in \cite{chen2013JSTSPpareto} that as long as the matrix $A$ is doubly-stochastic (i.e., satisfies $A\one=A^T\one=\one$) and $\mu$ is selected such that
	\begin{align}
		0	<	\mu		<	\min_{1 \le k \le N} \frac{1}{\sigma_{k}}
		\label{Equ:DictLearnDist:StepSizeCondition}
	\end{align}
where $\sigma_k$ is the Lipschitz constant\footnote{{
If $J_k(\nu; x_t)$ is twice-differentiable, then the Lipschitz gradient condition \eqref{Equ:DictLearnDist:LipschitzGradient} is  equivalent to requiring an upper bound on the Hessian of $J_k(\nu; x_t)$, i.e., 
		$0 	\le 	 \nabla_{\nu}^2 J_{k}(\nu; x_t)  	\le 	\sigma_{k} I_M$.
}} of the gradient of $J_k(\nu; x_t)$:
	\begin{align}
		\| \nabla_{\nu} J_{k}(\nu_1; x_t) - \nabla_{\nu} J_{k}(\nu_2; x_t) \| 
				\le 		\sigma_{k} \cdot \| \nu_1 - \nu_2 \|
		\label{Equ:DictLearnDist:LipschitzGradient}
	\end{align}
then algorithm \eqref{Equ:DictLearnDist:ATC_adapt}--\eqref{Equ:DictLearnDist:ATC_combine} converges to a fixed point that is $O(\mu^2)$ away from the optimal solution of \eqref{Equ:DictLearnDist:DualProblem_Objective_newForm} in squared Euclidean distance. We remark that a doubly-stochastic matrix is one that satisfies $A\one=A^T\one=\one$.

\begin{table*}[!ht]
\caption{Conjugate functions used in this paper for different tasks}
\label{Tab:ConjProx}
\centering
\renewcommand{\arraystretch}{2.0}

\begin{threeparttable}

\begin{tabular}{c||c|c|c|c|c|c|c|c}
\hline \hline
\rowcolor[gray]{0.9} \rule[-1ex]{0pt}{4ex} \textbf{Tasks} & $f(u)$ & $f^{\star}(\nu)$ & $\mc{V}_f$ & $ z_t^o$ & $h_{y_k}(y_k)$ & $h_{y_k}^{\star}(W_k^T \nu)$ & $\mc{V}_{h_{y_k}}$   &  $y_{k,t}^o$\\ 

\hline
\rule[-1ex]{0pt}{4ex} \textbf{Sparse SVD} & $\frac{1}{2} \|u\|_2^2$ & $\frac{1}{2}\|\nu\|_2^2$ & $\mb{R}^M$ & $x_t - \nu_t^o$ & $\gamma \|y_k\|_{1} + \frac{\delta}{2} \|y_k\|_2^2$ & $\mS_{\frac{\gamma}{\delta}}\left(\frac{W_k^T \nu}{\delta}\right)$ \tnote{b} & $\mb{R}^M$ & $\mT_{\frac{\gamma}{\delta}}\left(\frac{W_k^T \nu_t^o}{\delta}\right)$\tnote{a}  \\ 

\hline 
\rule[-1ex]{0pt}{4ex} \textbf{Bi-Clustering} & $\frac{1}{2}\|u\|_2^2$ & $\frac{1}{2}\|\nu\|_2^2$ & $\mb{R}^M$ & $x_t - \nu_t^o$ & $\gamma \|y_k\|_{1} + \frac{\delta}{2} \|y_k\|_2^2$ & $\mS_{\frac{\gamma}{\delta}}\left(\frac{W_k^T \nu}{\delta}\right)$ & $\mb{R}^M$ & $\mT_{\frac{\gamma}{\delta}}\left(\frac{W_k^T \nu_t^o}{\delta}\right)$ \\ 

\hline 
 \rule[0ex]{0pt}{4ex}\multirow{1}{*}{\textbf{Nonnegative Matrix}} & $\frac{1}{2}\|u\|_2^2$ & $\frac{1}{2}\|\nu\|_2^2$  & $\mb{R}^M$ & $x_t - \nu_t^o$ & $\gamma \|y_k\|_{1,+} + \frac{\delta}{2} \|y_k\|_2^2$ & $\mS_{\frac{\gamma}{\delta}}^{+}\left(\frac{W_k^T \nu}{\delta}\right)$ \tnote{d} & $\mb{R}^M$ & $\mT_{\frac{\gamma}{\delta}}^{+}\left(\frac{W_k^T \nu_t^o}{\delta}\right)$\tnote{c}   \\

 \cline{2-9} 
 \multirow{1}{*}{\textbf{Factorization}} & $\displaystyle\sum_{m=1}^M L(u_m)$ & $\frac{\eta}{2}\|\nu\|_2^2$  & $\{\nu: \|\nu\|_{\infty} \le 1\}$ & \cancel{\phantom{Nothing}} & $\gamma \|y_k\|_{1,+} + \frac{\delta}{2} \|y_k\|_2^2$ & $\mS_{\frac{\gamma}{\delta}}^{+}\left(\frac{W_k^T \nu}{\delta}\right)$ & $\mb{R}^M$  & $\mT_{\frac{\gamma}{\delta}}^{+}\left(\frac{W_k^T \nu_t^o}{\delta}\right)$ \\ 

\hline \hline
\end{tabular} 

\begin{tablenotes}

		\vspace{0.5em}
	\item[a]
		$\mT_{\lambda}(x)$ denotes the entry-wise soft-thresholding operator on
		the vector $x$: $[\mT_{\lambda}(x)]_n \defeq 
		(| [x]_n |-\lambda)_{+} \mathrm{sgn}([x]_n)$, where 
		$(x)_{+} = \max( x, 0 )$.
	
		\vspace{0.5em}
	\item[b]
		$\mS_{\frac{\gamma}{\delta}}(x)$ is the function defined by
		$\mS_{\frac{\gamma}{\delta}}(x)	\defeq
							-\frac{\delta}{2}
							\cdot
							\big\| \mT_{\frac{\gamma}{\delta}} 
							(x)
							\big\|_2^2
							-
							\gamma
							\cdot
							\big\|\mT_{\frac{\gamma}{\delta}}
							(x)
							\big\|_1
							+
							\delta \cdot x^T							
							\mT_{\frac{\gamma}{\delta}}
							\left(
								x
							\right)$
		for $x \in \mb{R}^M$.
	
		\vspace{0.5em}
	
	\item[c]
		$\mT_{\lambda}^{+}(x)$ denotes the entry-wise one-side
		soft-thresholding operator on the vector $x$: 
		$[\mT_{\lambda}^{+}(x)]_n \defeq ([x]_n - \lambda)_{+}$.
	
		\vspace{0.5em}
		
	\item[d]
		$\mS_{\frac{\gamma}{\delta}}^{+}(x)$ is defined by
		$\mS_{\frac{\gamma}{\delta}}^{+}(x)	\defeq
							-\frac{\delta}{2}
							\cdot
							\big\| \mT_{\frac{\gamma}{\delta}}^{+}(x)
							\big\|_2^2
							-
							\gamma
							\cdot
							\big\|\mT_{\frac{\gamma}{\delta}}^{+}
							(x)
							\big\|_1
							+
							\delta \cdot x^T							
							\mT_{\frac{\gamma}{\delta}}^{+}
							\left(
								x
							\right)$
		for $x \in \mb{R}^M$.
	
		\vspace{0.5em}


	\item[e]
		The functions $\mT_{\lambda}(x)$, $\mT_{\lambda}^{+}(x)$, 
		$\mS_{\frac{\gamma}{\delta}}(x)$, and 
		$\mS_{\frac{\gamma}{\delta}}^{+}(x)$ for the case of a scalar argument $x$ are illustrated in Fig.
		\ref{Fig:soft_thres_S_func}.
\end{tablenotes}

\end{threeparttable}

\end{table*}

%
%
%

Consider now the case in which the constraint set $\mc{V}_f$ is \emph{not} equal to $\mb{R}^{M}$ but is still known to all agents. This is a reasonable requirement. In general, we need to solve the supremum in \eqref{Equ:DictLearnDist:r_conj_def} with $r(x)=f(x)$ to derive the expression for $f^{\star}(\nu)$ and determine the set $\mc{V}_f$ that makes the supremum in \eqref{Equ:DictLearnDist:r_conj_def} finite. Fortunately, this step can be pursued in closed-form for many typical choices of $f(u)$. We list in Table \ref{Tab:ConjProx} the results that will be used in Sec. \ref{Sec:Experiments}; part of these results are derived in Appendix \ref{Appendix:DerivationConjugateFun} and the rest is from \cite[pp.90-95]{boyd2004convex}.  Usually, $\mc{V}_f$ for these typical choices of $f(u)$ are simple sets whose projection operators\footnote{The projection operator onto the set $\mc{V}_f$ is defined as $\displaystyle \Pi_{\mc{V}_f}(\nu)	\defeq	\arg\min_{x \in \mc{V}_f} \| x - \nu \|_2$.} can be found in closed-form --- see also \cite{parikh2013proximal}.
For example, the projection operator onto the set 
	\begin{align}
		\mc{V}_f 	= 	\{\nu: \| \nu \|_{\infty} \le 1\}
				=	\{ \nu: - \one \preceq \nu \preceq \one \}
	\end{align}
that is listed in the third row of Table \ref{Tab:ConjProx} is given by
	\begin{align}
		[\Pi_{ \mc{V}_f }( \nu)]_m
				=	\begin{cases}
						1		&		\mathrm{if~} \nu_m > 1		\\
						\nu_m	&		\mathrm{if~} -1 \le \nu_m \le 1			\\
						-1		&		\mathrm{if~} \nu_m < -1
					\end{cases} \label{Equ:DictLearnDist:Projection_Inf_Norm}
	\end{align}
where $[x]_m$ denotes the $m$-th entry of the vector $x$ and $\nu_m$ denotes the $m$-th entry of the vector $\nu$. Once the constraint set $\mc{V}_f$ is found, it can be enforced either by incorporating local projections onto $\mc{V}_f$ into the combination step \eqref{Equ:DictLearnDist:ATC_combine} at each agent \cite{theodoridis2011adaptive} or by using the penalized diffusion method \cite{towfic2013adaptive2}. For example, the projection-based strategy replaces \eqref{Equ:DictLearnDist:ATC_adapt}--\eqref{Equ:DictLearnDist:ATC_combine} by:
	\begin{subequations}
		\begin{align}
			\psi_{k,i}		&=		\nu_{k,i-1} - \mu \cdot \nabla_{\nu} J_k( \nu_{k,i-1}; x_t )
			\label{Equ:DictLearnDist:ATC_adapt_projection}
									\\
			\nu_{k,i}		&=		\Pi_{\mc{V}_f}
								\left[
									\sum_{ \ell \in \mN_k } a_{\ell k}  \psi_{\ell,i}
								\right]
			\label{Equ:DictLearnDist:ATC_combine_projection}
		\end{align}
	\end{subequations}
where $\Pi_{\mc{V}_f}[\cdot]$ is the projection operator onto $\mc{V}_f$. 

{
\subsection{Inference over Distributed Models: ADMM Strategies}
\label{Sec:DictLearn:Inference_ADMM}

An alternative approach to solving the dual inference problem \eqref{Equ:DictLearnDist:DualProblem_Objective_newForm}--\eqref{Equ:DictLearnDist:DualProblem_Constraint_newForm} is the distributed alternating direction multiplier method (ADMM) \cite{mota2012distributedBP,mota2013DADMM,Barbarossa2014distributed,schizas2008consensus1,zhu2009distributed}. Depending on the configuration of the network, there are different variations of distributed ADMM strategies. For example, the method proposed in \cite{schizas2008consensus1} relies on a set of bridge nodes for the distributed interactions among agents, and the method in \cite{mota2012distributedBP,mota2013DADMM} uses a graph coloring approach to partition the agents in the network into different groups, and lets the optimization process alternate between different groups with one group of agents engaged at a time. In \cite{zhu2009distributed} and \cite{Barbarossa2014distributed}, the authors developed ADMM strategies that adopt Jacobian style updates with all agents engaged in the computation concurrently. Below, we describe the Jacobian-ADMM strategies from\cite[p.356]{Barbarossa2014distributed} and briefly compare them with the diffusion strategies.

The Jacobian-ADMM strategy solves \eqref{Equ:DictLearnDist:DualProblem_Objective_newForm}--\eqref{Equ:DictLearnDist:DualProblem_Constraint_newForm} by first transforming it into the following equivalent optimization problem:
	\begin{subequations}
		\begin{align}
			\min_{\nu}		\quad&		\sum_{k=1}^N \big[ J_k(\nu_k; x_t) + I_{\mc{V}_f}(\nu_k) \big]
			\label{Equ:DictLearnDist:DualProblem_Objective_newForm_ADMM}
									\\
			\mathrm{s.t.}	\quad&		\nu_k = \nu_{\ell}, \quad \ell \in \mc{N}_k\backslash\{k\}, 
									\;\; k=1,\ldots,N
			\label{Equ:DictLearnDist:DualProblem_Constraint_newForm_ADMM}
		\end{align}
	\end{subequations}
where the cost function is decoupled among different $\{\nu_k\}$ and the constraints are coupled through neighborhoods. Then, the following recursion is used to solve \eqref{Equ:DictLearnDist:DualProblem_Objective_newForm_ADMM}--\eqref{Equ:DictLearnDist:DualProblem_Constraint_newForm_ADMM}:
	\begin{subequations}
	\begin{align}
		\nu_{k,i}			&=		\arg\min_{\nu_k}
								\sum_{k=1}^N 
								\bigg\{
									\big[ J_k(\nu_k; x_t) + I_{\mc{V}_f}(\nu_k) \big]
									\nn\\
									&\quad+ \!
									\sum_{\ell=1}^N 
									\! b_{k \ell} 
									\Big[ \!
									\lambda_{k \ell,i\!-\!1} ^T
									( \nu_{\ell,i \!-\! 1} \!-\! \nu_{k} )
									\!+\!
									\| \nu_{\ell,i-1} \!-\! \nu_{k} \|_2^2
									\Big] \!
								\bigg\}
		\label{Equ:DictLearnDist:ADMM_primal}
								\\
		\lambda_{k\ell,i} \!	&=		\! \lambda_{k\ell,i-1} 
								+ 
								\mu \; b_{k \ell} \cdot
								\left(
									\nu_{k,i} - \nu_{\ell,i}
								\right)
		\label{Equ:DictLearnDist:ADMM_dual}
	\end{align}
	\end{subequations}
where $b_{k\ell}$ is the $(k, \ell)$-th entry of the adjacency matrix $B = [b_{k\ell}]$ of the network, which is defined as:
	\begin{align}
		b_{k\ell}	=	1  \mathrm{~if~} \ell \in \mc{N}_k \backslash \{k\},
		\; b_{k \ell}=0 \mathrm{~otherwise}
	\end{align}	
	\begin{figure}[t!]
		\centering
		\includegraphics[width=0.48\textwidth]{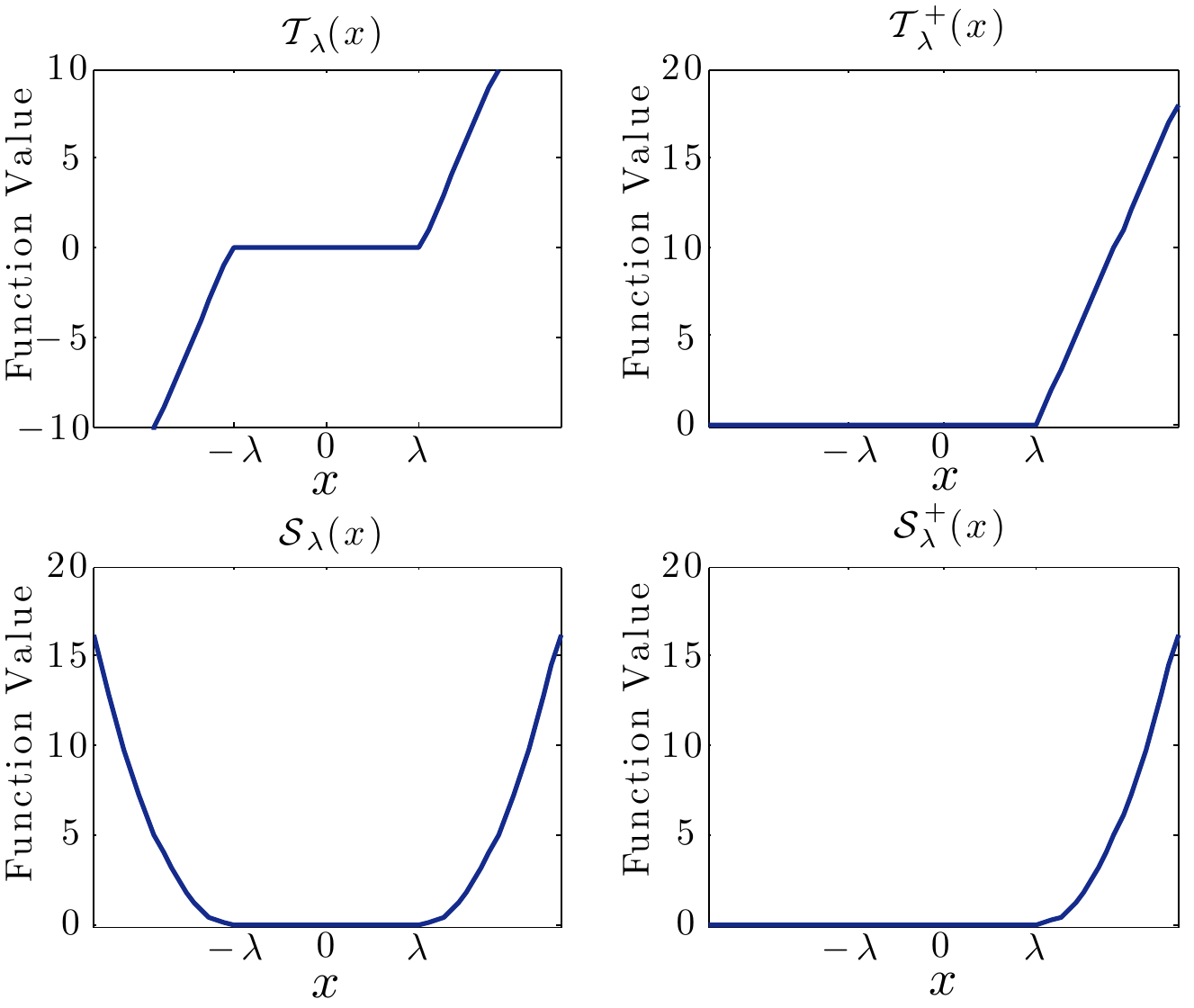}
		\caption{Illustration of the functions $\mT_{\lambda}(x)$, $\mT_{\lambda}^+(x)$, $\mS_{\lambda}(x)$, and $\mS_{\lambda}^+(x)$.}
		\label{Fig:soft_thres_S_func}
	\end{figure}%
	\begin{figure}[t]
		\centering
		\includegraphics[width=0.45\textwidth]{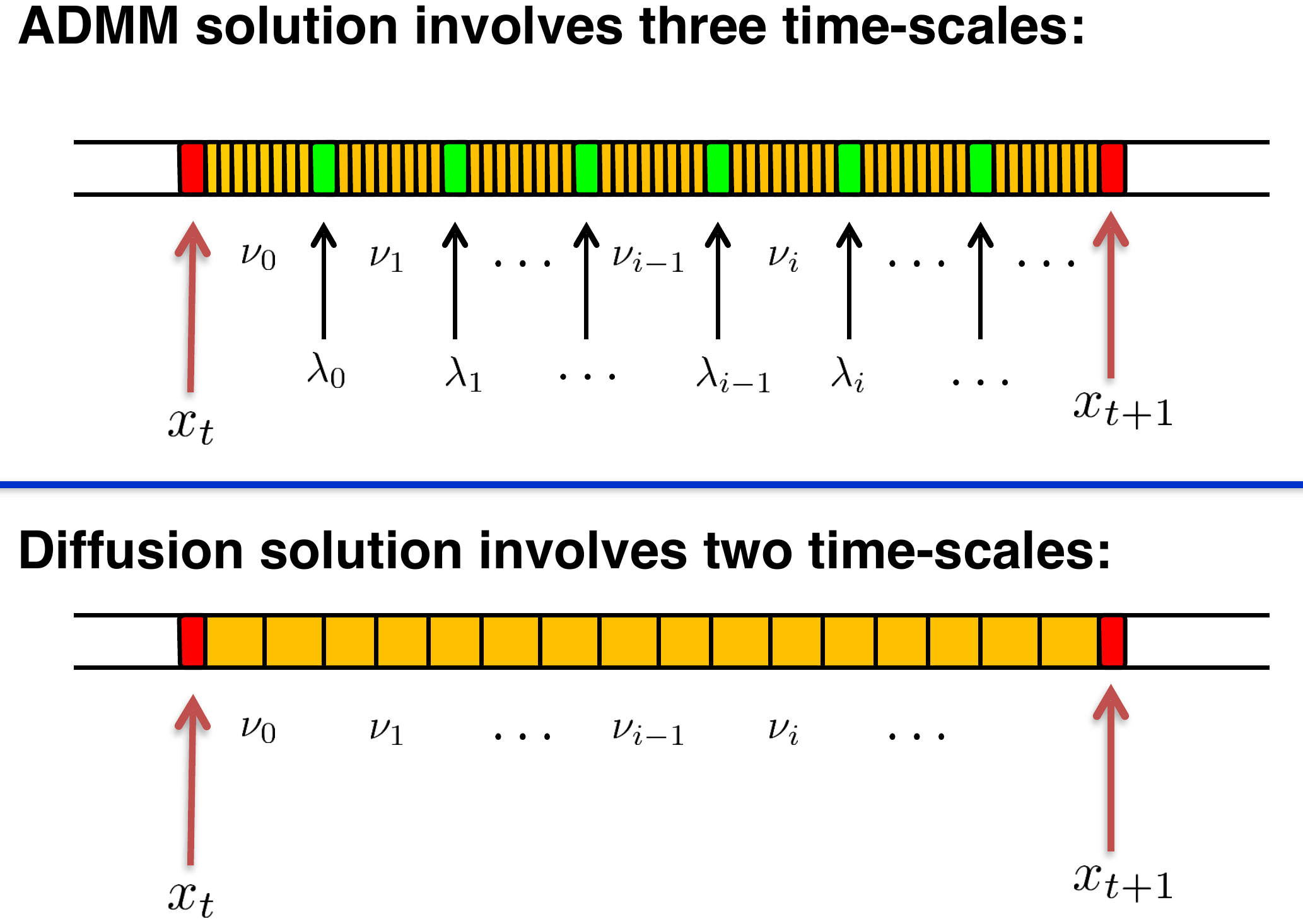}
		\caption{Comparison between the ADMM strategy and the diffusion strategy. The diffusion strategy
		has two time scales and the ADMM strategy may have three time scales. The first time scale is 
		the dictionary update over the data stream (see Sec. \ref{Sec:DictLearn:DictUpdate}), 
		the second time scale is the iterative algorithm for solving the inference problem for each data
		sample $x_t$, and the third time scale in ADMM is to solve the ``argmin'' in 
		\eqref{Equ:DictLearnDist:ADMM_primal}.}
		\label{Fig:ADMMvsDiffusion}
	\end{figure}%
From recursion \eqref{Equ:DictLearnDist:ADMM_primal}--\eqref{Equ:DictLearnDist:ADMM_dual}, we observe that ADMM requires solving a separate optimization problem ($\arg\min$) for each ADMM step. This optimization problem generally requires an iterative algorithm to solve when it cannot be solved in closed-form, which adds a third time scale to the algorithm, as explained in\cite{towfic2014dictionary} in the context of dictionary learning. This situation is illustrated in Fig. \ref{Fig:ADMMvsDiffusion}. The need for a third time-scale usually translates into requiring faster processing at the agents between data arrivals, which can be a hindrance for adaptation in real-time.

}

\subsection{Recovery of the Primal Variables}
\label{Sec:DictLearn:RecoveryPrimalVar}

Returning to the diffusion solution \eqref{Equ:DictLearnDist:ATC_adapt}--\eqref{Equ:DictLearnDist:ATC_combine} or \eqref{Equ:DictLearnDist:ATC_adapt_projection}--\eqref{Equ:DictLearnDist:ATC_adapt_projection}, once the optimal dual variable $\nu_t^o$ has been estimated by the various agents, the optimal primal variables $y_{k,t}^o$ and $z_t^o$ can now be recovered uniquely if $f(u)$ and $\{h_{y_k}(y_k)\}$ are strongly convex. In this case, the infimums in \eqref{Equ:DictLearnDist:DualFunction_def} can be attained and become minima. As a result, optimal primal variables can be recovered via
	\begin{align}
		z_t^o		&=			\arg\min_{ z }
									\left\{
										f( x_t - z ) + (\nu_t^o)^T z
									\right\}
									\nn\\
					&\overset{(a)}{=}
									x_t - \arg\max_{u} \big[ (\nu_t^o)^T u - f(u) \big]
		\label{Equ:DictLearnDist:z_optimal_primal}
									\\
		y_{k,t}^o	&=
									\arg\min_{y_k}
										\Big\{
											h_{y_k}(y_k) 
											\!-\!
											(\nu_t^o)^T W_k y_k
										\Big\}
									\nn\\
				&=
									\arg\max_{y_k} 
									\big[
										(W_k^T \nu_t^o)^T y_k - h_{y_k}(y_k)
									\big]
		\label{Equ:DictLearnDist:yk_optimal_primal}
	\end{align}
where step (a) performs the variable substitution $u=x_t-z$.
By \eqref{Equ:DictLearnDist:z_optimal_primal}--\eqref{Equ:DictLearnDist:yk_optimal_primal}, we obtain the optimal solutions of \eqref{Equ:DictLearnDist:NewInference_Objective}--\eqref{Equ:DictLearnDist:NewInference_Constraint} (and also of the original inference problem \eqref{Equ:ProbForm:InferenceProblem}) after first solving the dual problem \eqref{Equ:DictLearnDist:DualProblem}. For many typical choices of $f(\cdot)$ and $h_{y_k}(\cdot)$, the solutions of \eqref{Equ:DictLearnDist:z_optimal_primal}--\eqref{Equ:DictLearnDist:yk_optimal_primal} can be expressed in closed form in terms of $\nu_t^o$. In Table \ref{Tab:ConjProx}, we list the results that will be used later in Sec. \ref{Sec:Experiments} with the derivation given in Appendix \ref{Appendix:DerivationConjugateFun}. 

The strong convexity of $f(u)$ and $\{h_{y_k}(y_k)\}$ is needed if we want to uniquely recover $z_t^o$ and $\{y_{k,t}^o\}$ from the dual problem \eqref{Equ:DictLearnDist:DualProblem}. As we will show further ahead in \eqref{Equ:DictLearnDist:DictionaryUpdate_final}, the quantities $\{y_{k,t}^o\}$ are always needed in the dictionary update. For this reason, we assumed in Assumption \ref{Assumption:StronglyConvexRegularization_hyk} that the $\{h_{y_k}(y_k)\}$ are strongly convex, which can always be satisfied by means of elastic net regularization as explained earlier. On the other hand, depending on the application, the recovery of $z_t^o$ is not always needed and neither is the strong convexity of $f(u)$ (in these cases, it is sufficient to assume that $f(u)$) is convex). For example, as explained in \cite{chen2014icasspdictionary}, the image denoising application requires recovery of $z_t^o$ as the final reconstructed image. On the other hand, the novel document detection application discussed further ahead does not require recovery of $z_t^o$ but the maximum value of the dual function, $g(\nu; x_t)$, which, by strong duality, is equal to the minimum value of the cost function \eqref{Equ:DictLearnDist:NewInference_Objective} and that of \eqref{Equ:ProbForm:InferenceProblem}.

\subsection{Choice of Residual and Regularization Functions}
\label{Sec:DictLearn:ChoiceResReg}
In Tables \ref{Tab:Task}--\ref{Tab:ConjProx}, we list several typical choices for the residual function, $f(u)$, and the regularization functions, $\{h_{y_k}(y_k)\}$. In general, a careful choice of $f(u)$ and $\{h_{y_k}(y_k)\}$ can make the dual cost \eqref{Equ:DictLearn:Inference_Dual_Objective} better conditioned than in the primal cost \eqref{Equ:DictLearnDist:NewInference_Objective}. Recall that the primal cost \eqref{Equ:DictLearnDist:NewInference_Objective} may not be differentiable due to the choice of $h_{y_k}(y_k)$ (e.g., the elastic net). However, if $f(u)$ is chosen to be strictly convex with Lipschitz gradients and the $\{h_{y_k}(y_k)\}$ are chosen to be strongly convex (not necessarily differentiable), then the conjugate function $f^{\star}(\cdot)$ will be a differentiable strongly convex function with Lipschitz gradient and the $\{h_{y_k}^{\star}(\cdot)\}$ will be differentiable convex functions with Lipschitz gradients \cite[pp.79--84]{urruty1993convex2}. Adding $f^{\star}(\cdot)$ and $\{h_{y_k}^{\star}(\cdot)\}$  together in \eqref{Equ:DictLearn:Inference_Dual_Objective} essentially transforms a non-differentiable primal cost \eqref{Equ:DictLearnDist:NewInference_Objective} into a differentiable strongly convex dual cost  \eqref{Equ:DictLearn:Inference_Dual_Objective} with Lipschitz gradients. As a result, the algorithms that optimize the dual problem \eqref{Equ:DictLearn:Inference_Dual_Objective}--\eqref{Equ:DictLearn:Inference_Dual_Constraint} can generally enjoy a fast (geometric) convergence rate \cite{poliak1987introduction,chen2013JSTSPpareto,sayed2014adaptation}.

\subsection{Distributed Dictionary Updates}
\label{Sec:DictLearn:DictUpdate}
Now that we have shown how the inference task \eqref{Equ:ProbForm:InferenceProblem} can be solved in a distributed manner, we move on to explain how the local sub-dictionaries $W_k$ can be updated through the solution of the stochastic optimization problem \eqref{Equ:ProbForm:DictLearn_Objective}--\eqref{Equ:ProbForm:DictLearn_Constraint}, which is rewritten as:
	\begin{subequations}
	\label{Equ:DictLearnDist:DictUpdate_total}
		\begin{align}
			\min_{W}	\quad&		\Expt
									Q(W, \y_t^o; \x_t)
									+
									\sum_{k=1}^N
									h_{W_k}(W_k)
			\label{Equ:DictLearnDist:DictUpdate_Objective}
									\\
			\mathrm{s.t.}
						\quad&		
									W_k \in \mW_k,
									\quad
									k = 1,\ldots, N
			\label{Equ:DictLearnDist:DictUpdate_Constraint}
		\end{align}
	\end{subequations}
where the loss function $Q(W, \y_t^o; \x_t)$ is given in \eqref{Equ:ProbForm:InferenceProblem_FactoredForm},  $\y_t^o \defeq \col\{ \y_{1,t}^o,\ldots,\y_{N,t}^o \}$, the decomposition for $h_W(W)$ from \eqref{Equ:ProbForm:hW_factorization} is used, and we assume the constraint set $\mW$ can be decomposed into a set of constraints $\{\mW_k\}$ on the individual sub-dictionaries $W_k$; this condition usually holds for typical dictionary learning applications --- see Table \ref{Tab:Task}. 
{
Problem \eqref{Equ:DictLearnDist:DictUpdate_Objective}--\eqref{Equ:DictLearnDist:DictUpdate_Constraint} can also be written as the following unconstrained optimization problem by introducing indicator functions for the sets $\{\mc{W}_k\}$:
		\begin{align}
			\min_{W}	\quad&		\Expt
									Q(W, \y_t^o; \x_t)
									\!+\!
									\sum_{k=1}^N
									\Big[
										h_{W_k}(W_k)
										+
										I_{\mc{W}_k}(W_k)
									\Big]
			\label{Equ:DictLearnDist:DictUpdate_Objective_Unconstrained}
		\end{align}
}%
Note that the cost function in \eqref{Equ:DictLearnDist:DictUpdate_Objective_Unconstrained} consists of two parts, where the first term is differentiable\footnote{Note from \eqref{Equ:ProbForm:InferenceProblem_FactoredForm} that $Q(\cdot)$ depends on $W$ via $f(\cdot)$, which is assumed to be differentiable.}  with respect to $W$ while the second term, if it exists, is non-differentiable but usually consists of simple components --- see Table \ref{Tab:Task}. A typical approach to optimizing cost functions of this type is the \emph{proximal gradient} method\cite{figueiredo2005bound,figueiredo2007majorization,beck2009fast,parikh2013proximal}, which applies gradient descent to the first differentiable part followed by a proximal operator to the second non-differentiable part. This method is known to converge faster than applying the subgradient descent method to both parts.
{ However, the proximal gradient methods in\cite{figueiredo2005bound,figueiredo2007majorization,beck2009fast,parikh2013proximal} are developed for deterministic optimization, where the exact form of the objective function is known. In constrast, our objective function in \eqref{Equ:DictLearnDist:DictUpdate_Objective_Unconstrained} assumes a stochastic form and is unknown beforehand because the statistical distribution of the data $\{\x_t\}$ is not known.
}%
Therefore,  our strategy is to apply the proximal gradient method to the cost function in \eqref{Equ:DictLearnDist:DictUpdate_Objective_Unconstrained} and remove the expectation operator to obtain an instantaneous approximation to the true gradient; this is the approach typically used in adaptation \cite{Sayed08,sayed2014adaptation,sayed2014proc} and stochastic approximation\cite{kushner2003stochastic}:
	\begin{align}
		W_{k,t}		\!=\!		\mathrm{prox}_{\mu_w  \cdot  (h_{W_k}\!+\!I_{\mc{W}_k})}
								\Big\{
									W_{k,t-1} \!-\! \mu_w \nabla_{W_k} Q(W_{t-1}, y_t^o; x_t)
								\Big\}
		\label{Equ:DictLearn:DictUpdate_interm1_alternative}
	\end{align}
Recursion \eqref{Equ:DictLearn:DictUpdate_interm1_alternative} is effective as long as the proximal operator of $h_{W_k}(W_k)+I_{\mc{W}_k}(W_k)$ can be solved easily in closed-form. When this is not possible but the proximal operators of $h_{W_k}(\cdot)$ and $I_{\mc{W}_k}(\cdot)$ are simple, it is preferable to apply a stochastic gradient descent step, followed by the proximal operator of $h_{W_k}(\cdot)$, and then the proximal operator of $I_{\mc{W}_k}(\cdot)$ (equivalent to $\Pi_{\mc{W}_k}(\cdot)$\cite{parikh2013proximal}, which is the projection onto $\mc{W}_k$) in an \emph{incremental} manner \cite{bertsekas2010incremental}, thus leading to the following recursion:
	\begin{align}
		W_{k,t}		\!=\!		\Pi_{\mW_k}\!\!
							\left\{\!
								\mathrm{prox}_{\mu_w \!\cdot h_{W_k}}
								\!\!
								\big(
									W_{k,t-1} \!-\! \mu_w \nabla_{W_k} Q(W_{t-1}, y_t^o; x_t)
								\big)\!
							\right\}
		\label{Equ:DictLearn:DictUpdate_interm1}
	\end{align}
where $W_{t-1} \defeq [ W_{1,t-1}, \cdots, W_{N,t-1} ]$, and $\mathrm{prox}_{\mu_w \cdot h_{W_k}}(\cdot)$ denotes the proximal operator of $\mu_w \cdot h_{W_k}(W_k)$. The expression for the gradient $\mu_w \nabla_{W_k} Q(W_{t-1}, y_t^o; x_t)$ will be given further ahead in \eqref{Equ:DictLearn:DictUpdate_gradWk_interm1}--\eqref{Equ:DictLearnDist:DictionaryUpdate_final}. We recall that the proximal operator of a vector function $h(u)$ is defined as \cite[p.6]{parikh2013proximal}:
	\begin{align}
		\mathrm{prox}_{h}(x)		
							\defeq	
									\arg\min_{u} 
									\left(
										h(u) + \frac{1}{2} \| u - x \|_2^2
									\right)
		\label{Equ:DictLearnDist:Prox_def}
	\end{align}	
For a matrix function $h(U)$, the proximal operator assumes the same form as \eqref{Equ:DictLearnDist:Prox_def} except that the Euclidean norm in \eqref{Equ:DictLearnDist:Prox_def} is replaced by the Frobenius norm.  The proximal operator for $\mu_w \cdot h_{W_k}(W_k) = \mu_w\beta \cdot \vertiii{W_k}_1$ used in the bi-clustering task in Table \ref{Tab:Task} is the entry-wise soft-thresholding function \cite[p.191]{parikh2013proximal}:
	\begin{align}
		\mathrm{prox}_{\mu_w \cdot h_{W_k}}(\cdot)
					=
							\mathrm{prox}_{\mu_w \beta \cdot \vertiii{W_k}}(\cdot)
					=		
							\mc{T}_{\mu_w \cdot \beta}(\cdot)
		\label{Equ:DictLearn:DictUpdate_ProxW_SoftThresh}
	\end{align}
and the proximal operator for $h_{W_k}(W_k) = 0$ for other cases in Table \ref{Tab:Task} is the identity mapping: $\mathrm{prox}_{0}(x)	=x$. With regards to the projection operator used in \eqref{Equ:DictLearn:DictUpdate_interm1}, we provide some examples of interest for the current work. If the constraint set $\mW_k$ is of the form:
	\begin{align}
		\mW_k = \{ W_k: \; \|[W_k]_{:,q}\|_2 \le 1\}
	\end{align} 
then the projection operator $\Pi_{\mW_k}(\cdot)$ is given by\cite{theodoridis2011adaptive,parikh2013proximal}:
	\begin{align}
		[\Pi_{\mW_k}(X)]_{:,n}	
					=		\begin{cases}
								[X]_{:,n},						&	\|[X]_{:,n}\|_2 \le 1		\\
								\frac{[X]_{:,n}}{\|[X]_{:,n}\|_2},		&	\|[X]_{:,n}\|_2 > 1
							\end{cases}
		\label{Equ:DictLearn:DictUpdate_Projection_normball}
	\end{align}
On the other hand, if the constraint set $\mW_k$ is of the form:
	\begin{align}
		\mW_k = \{ W_k: \; \|[W_k]_{:,q}\|_2 \le 1, \; \mW \succeq 0\}
	\end{align}
then the projection operator $\Pi_{\mW_k}(\cdot)$ becomes
	\begin{align}
		[\Pi_{\mW_k}(X)]_{:,n}	
					=		\begin{cases}
								\big([X]_{:,n}\big)_{+},						
										&	\|\big([X]_{:,n}\big)_{+}\|_2 \le 1
								\\
								\displaystyle
								\frac{\big([X]_{:,n}\big)_{+}}
								{\|\big([X]_{:,n}\big)_{+}\|_2},		
										&	\|\big([X]_{:,n}\big)_{+}\|_2 > 1
							\end{cases}
		\label{Equ:DictLearn:DictUpdate_Projection_nonnegativenormball}
	\end{align}
where $(x)_{+} = \max(x,0)$, i.e., it replaces all the negative entries of a vector $x$ with zeros.

\noindent\begin{algorithm}[t]
	\caption{{\small Model-distributed diffusion strategy for dictionary learning (Main algorithm)}}
	\label{alg:DiffusionSparseCoding}
	{\small{
	\begin{algorithmic}
	{
	\STATE {\bf Initialization:} The sub-dictionaries $\{W_k\}$ are randomly initialized and then projected onto either the constraint \eqref{Equ:ProbForm:W_subUnitNormConstraint} or \eqref{Equ:ProbForm:W_subUnitNormNonnegConstraint}, depending on the task in Tab. \ref{Tab:Task}.}
	\FOR{each input data sample $x_t$}
		\STATE Compute $\nu_{t}^o$ by iterating  \eqref{Equ:DictLearnDist:ATC_adapt}-\eqref{Equ:DictLearnDist:ATC_combine} until convergence: $\nu_t^o\approx \nu_{k,i}$. That is:
				\begin{align*}
					\begin{cases}
						\psi_{k,i}		=		\nu_{k,i-1} 
											- 
											\mu \cdot \nabla_{\nu} J_k( \nu_{k,i-1}; x_t )
												\\						
						\nu_{k,i}		=		\displaystyle
											\Pi_{\mathcal{V}_f}\Big\{\sum_{ \ell \in \mN_k } a_{\ell k}  \psi_{\ell,i}\Big\}
					\end{cases}
				\end{align*}
			{ with initialization $\{\nu_{k,0} = 0, \; k=1,\ldots,N\}$.}
		\FOR{each agent $k$}
			\STATE Compute coefficient $y_{k,t}^o$ using Table \ref{Tab:ConjProx} or 
			\eqref{Equ:DictLearnDist:yk_optimal_primal}:
				\begin{align*}
					y_{k,t}^o		=		\arg\max_{y_k} 
										\big[
											(W_k^T \nu_t^o)^T y_k - h_{y_k}(y_k)
										\big]
				\end{align*}
				\vspace{-0.8em}
			\STATE Adjust dictionary element $W_{k,t}$ using \eqref{Equ:DictLearnDist:DictionaryUpdate_final}:
				\begin{align*}
					W_{k,t}		=		\Pi_{\mW_k}
										\left\{
											\mathrm{prox}_{\mu_w \cdot h_{W_k}}
											\big(
												W_{k,t-1} + \mu_w \nu_t^o (y_{k,t}^o)^T
											\big)
										\right\}
				\end{align*}
		\ENDFOR
	\ENDFOR
	\end{algorithmic}}}
\end{algorithm}

\begin{figure}
	\centering
	\includegraphics[width=0.48\textwidth]{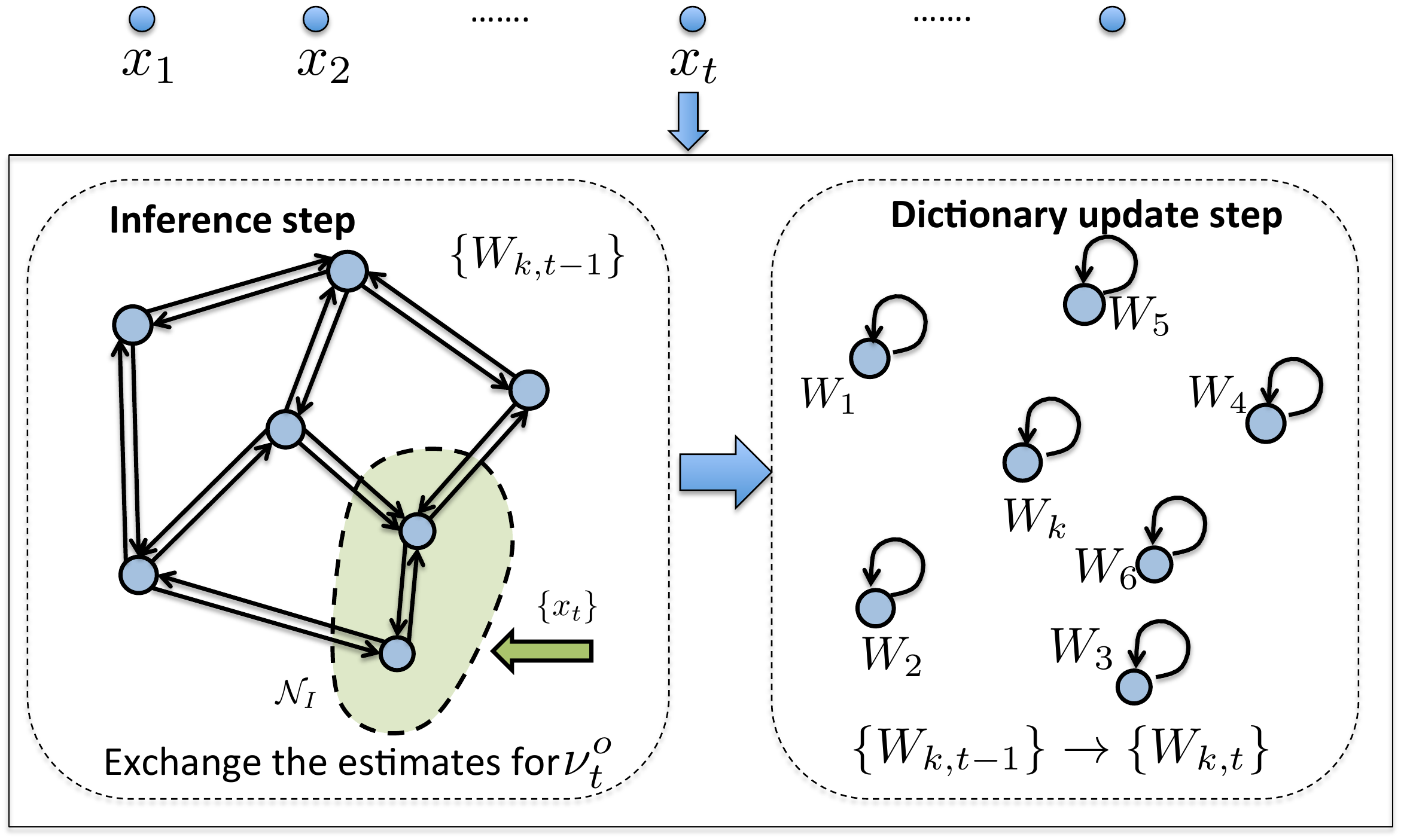}
	\caption{The distributed inference step and the dictionary update step over distributed models. 
	In the inference step, after each data sample $x_t$ arrives a subset of the agents in the network,
	all the agents find the corresponding optimal dual variable $\nu_t^o$ by exchanging the estimates of 
	$\nu_t^o$ with neighbors. In the dictionary update step, agents update their sub-dictionaries locally 
	on their own using a step of proximal stochastic gradient descent as
	\eqref{Equ:DictLearnDist:DictionaryUpdate_final}.}
	\label{Fig:Fig_DictLearn}
\end{figure}

Now, we return to derive the expression for the gradient $\nabla_{W_k} Q(W_{t-1}, y_t^o; x_t)$ in \eqref{Equ:DictLearn:DictUpdate_interm1}. By \eqref{Equ:ProbForm:InferenceProblem_FactoredForm}, we have 
	\begin{align}
		\nabla_{W_k} \! Q(W_{t-1}, y_t^o; x_t)
					\!&=		\!-\!
							f_{u}'
							\!
							\Big(\!
								x_t \!-\! \sum_{k=1}^N W_{k,t-1} y_{k,t}^o
								\!
							\Big)
							(y_{k,t}^o)^T \!\!
		\label{Equ:DictLearn:DictUpdate_gradWk_interm1}
	\end{align}
where $f_{u}'(u)$ denotes the gradient of $f(u)$ with respect to the residual $u$. On the face of it, expression \eqref{Equ:DictLearn:DictUpdate_gradWk_interm1} requires global knowledge by agent $k$ of all sub-dictionaries $\{W_k\}$ across the network, which goes against the desired objective of arriving at a distributed implementation. However, we can develop a distributed algorithm by exploiting the structure of the problem as follows. Note from \eqref{Equ:DictLearnDist:Lagrangian} that the optimal inference result should satisfy:
	\begin{align}
		\begin{cases}
			0		=			\nabla_z L  ( \{y_{k,t}^o\}, z_t^o, \nu_t^o; x_t ) 		
								\\
			0		=			\nabla_\nu L  ( \{y_{k,t}^o\}, z_t^o, \nu_t^o; x_t )
		\end{cases}
		\!\! \Leftrightarrow \;\;
		\begin{cases} \!
			0		=			- \! f_{u}'( x_t \!-\! z_t^o ) \!+\! \nu_t^o
								\\
			\displaystyle\!
			z_t^o	=			\sum_{k=1}^N
								W_{k,t-1} y_{k,t}^o
		\end{cases}\!\!\!\!\!\!\!\!\!
		\label{Equ:DictLearn:DictUpdate_OptimalInferenceCondition}
	\end{align}
which leads to
	\begin{align}
		&0 		= 		- f_{u}'\Big(
								x_t - \sum_{k=1}^N W_{k,t-1} y_{k,t}^o
							\Big)
							+
							\nu_t^o
							\nn\\
		&\quad\Leftrightarrow\quad
				\nu_t^o = 	f_{u}'\Big(
								x_t - \sum_{k=1}^N W_{k,t-1} y_{k,t}^o
							\Big)
		\label{Equ:Dictlearn:DictUpdate_grad_nu_relation}
	\end{align}
In other words, we find that  the optimal dual variable $\nu_t^o$ is equal to the desired gradient vector. Substituting \eqref{Equ:Dictlearn:DictUpdate_grad_nu_relation} into \eqref{Equ:DictLearn:DictUpdate_gradWk_interm1}, the dictionary learning update \eqref{Equ:DictLearn:DictUpdate_interm1} becomes
	\begin{align}
		W_{k,t}		=		\Pi_{\mW_k}
							\left\{
								\mathrm{prox}_{\mu_w \cdot h_{W_k}}
								\big(
									W_{k,t-1} + \mu_w \nu_t^o (y_{k,t}^o)^T
								\big)
							\right\}
		\label{Equ:DictLearnDist:DictionaryUpdate_final}
	\end{align}
which is now in a fully-distributed form. At each agent $k$, the above $\nu_t^o$ can be replaced by the estimate $\nu_{k,i}$ after a sufficient number of inference iterations (large enough $i$). We note that the dictionary learning update \eqref{Equ:DictLearnDist:DictionaryUpdate_final} has the following important interpretation. Let
	\begin{align}
		u_t^o		\defeq
							x_t - \sum_{k=1}^N W_{k,t-1} y_{k,t}^o
	\end{align}
which is the optimal prediction residual error using the entire existing dictionary set $\{W_{k,t-1}\}_{k=1}^N$. Observe from \eqref{Equ:Dictlearn:DictUpdate_grad_nu_relation} that $\nu_t^o$ is the gradient of the residual function $f(u)$ at the optimal $u_t^o$. The update term for dictionary element $k$ in \eqref{Equ:DictLearnDist:DictionaryUpdate_final} is effectively the correlation between $\nu_t^o$, the gradient of the residual function $f(u_t^o)$,  and the coefficient $y_{k,t}^o$ (the activation) at agent $k$. In the special case of $f(u) = \frac{1}{2} \|u\|_2^2$, expression \eqref{Equ:Dictlearn:DictUpdate_grad_nu_relation} implies that
	\begin{align}
		\nu_t^o 		=		u_t^o
					=
							x_t - \sum_{k=1}^N W_{k,t-1} y_{k,t}^o
	\end{align}
In this case, $\nu_t^o$ has the interpretation of being equal to the optimal prediction residual error, $u_t^o$, using the entire existing dictionary set $\{W_{k,t-1}\}_{k=1}^N$. Then, the update term for dictionary element $k$ in \eqref{Equ:DictLearnDist:DictionaryUpdate_final} becomes the correlation between the optimal prediction error $\nu_t^o = u_t^o$  and the coefficient $y_{k,t}^o$ at agent $k$. Furthermore, recursion \eqref{Equ:DictLearnDist:DictionaryUpdate_final} reveals that, for each input data sample $x_t$, after the dual variable $\nu_t^o$ is obtained at each agent, there is no need to exchange any additional information among agents in order to update their own sub-dictionaries; the dual variable $\nu_t^o$ already provides sufficient information to carry out the update. The fully distributed algorithm for dictionary learning is listed in Algorithm \ref{alg:DiffusionSparseCoding} and is also illustrated in Fig. \ref{Fig:Fig_DictLearn}.

\section{Important Special Cases and Experiments}
\label{Sec:Experiments}
In this section, we apply the dictionary learning algorithm to two problems involving novel document/topic detection and bi-clustering.  A third application to image denoising is considered in \cite{chen2014icasspdictionary,towfic2014dictionary}.\footnote{The software code for the experiments in this manuscript is available online at \url{http://www.ee.ucla.edu/asl}} In our experiments below, we will use the diffusion strategy \eqref{Equ:DictLearnDist:ATC_adapt}--\eqref{Equ:DictLearnDist:ATC_combine} or \eqref{Equ:DictLearnDist:ATC_adapt_projection}--\eqref{Equ:DictLearnDist:ATC_combine_projection} to solve the dual inference problem \eqref{Equ:DictLearn:Inference_Dual_Objective}--\eqref{Equ:DictLearn:Inference_Dual_Constraint}.


\subsection{Tuning of the parameters}
\label{Sec:Discussion}

	In the following experiments, it is necessary to select properly the step-size $\mu$ for the diffusion algorithm \eqref{Equ:DictLearnDist:ATC_adapt}--\eqref{Equ:DictLearnDist:ATC_combine} to ensure that the estimate for $\nu_t^o$ converges sufficiently close to it after a reasonable number of iterations. { Table \ref{Tab:StepSizeCond} lists the step-size conditions that guarantee the convergence of the diffusion algorithm for different applications, which are derived from the general condition \eqref{Equ:DictLearnDist:StepSizeCondition}. Note that as long as the agents know the regularization parameter $\delta$ and the maximum number,  $N_{\max}$, of dictionary atoms that are allowed at each agent, the agents can select the step-size in a distributed manner.}
		
\begin{figure}[t!]
	\centering
	\includegraphics[width=0.45\textwidth]{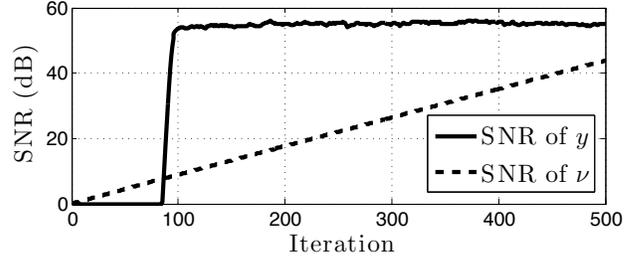}
	\caption{Learning curve for the Huber document detection example described by Alg.~\ref{alg:DocumentDetection:Huber} with $\mu = 0.5$.}	
	\label{Fig:Discussion:LearningCurve}
\end{figure}

	{ For the convenience of the experiments in this section and only to get an idea about how many iterations are typically needed for the inference step, we choose a data sample $x$ from the training dataset, and use a non-distributed optimization package such as CVX \cite{cvx} to compute the optimal solution $y^o \triangleq \col\{y_{1}^o,\ldots, y_N^o\}$ and its respective dual variable $\nu^o$ as the ground truth for the inference problem \eqref{Equ:DictLearnDist:NewInference_Objective}--\eqref{Equ:DictLearnDist:NewInference_Constraint}. We plot the signal-to-noise measures $\|y^o\|^2/\|y_i-y^o\|^2$ and $\|\nu^o\|^2/\|\nu_{k,i}-\nu^o\|^2$ against the iteration number $i$ in Fig.~\ref{Fig:Discussion:LearningCurve}. The value $\nu_{k,i}$ is obtained from the distributed algorithm (see \eqref{Equ:DictLearnDist:ATC_combine} or \eqref{Equ:DictLearnDist:ATC_combine_projection}) at each iteration $i$ and $y_i \triangleq \col\{y_{1,i},\ldots,y_{N,i}\}$ is calculated at each iteration according to:
	\begin{align}
		y_{k,i}		=		\arg\max_{y_k} 
							\big[
								(W_k^T \nu_{k,i})^T y_k - h_{y_k}(y_k)
							\big]
		\label{Equ:DictLearnDist:yk_optimal_primal_tuning_section}
	\end{align}	
Observe from Fig. \ref{Fig:Discussion:LearningCurve} that in order to achieve satisfactory SNR values (e.g., $40$-$50$dB) for both $y$ and $\nu$, the required number of diffusion iterations is about $500$. Also note that the primal variable $y$ generally reaches a high SNR value before the dual variable $\nu$, but both are required to be found with reasonable accuracy for the dictionary update step (see \eqref{Equ:DictLearnDist:DictionaryUpdate_final}). Furthermore, although the number of iterations by diffusion seems to be large for solving the inference problem, the actual wall-clock time it takes is short because of the relatively low complexity per step. 

We further note that there was no restriction imposed on the size of the network. In our experiments, network consists of $N=196$ nodes in the image denoising example \cite{chen2014icasspdictionary} and employs from $N=10$ to $N=80$ nodes in  the novel document detection example. In the bi-clustering example, the network size, $N$, is three because of the application setup and the nature of the data from \cite{lee2010biclustering}, where the rank of the data matrix is low so that three dictionary atoms are sufficient to represent the data. 
}

\begin{table}[!t]
\caption{Conditions of the step-size parameter $\mu$ for inference step.}
\label{Tab:StepSizeCond}
\centering
\renewcommand{\arraystretch}{2.0}

\begin{threeparttable}

\begin{tabular}{c|c||c}
\hline \hline
\rowcolor[gray]{0.9} \rule[-1ex]{0pt}{4ex}  $f(u)$ & $h_{y_k}(y_k)$ &  Step-size condition\\

\hline 
\rule[-1ex]{0pt}{4ex} $\frac{1}{2}\|u\|_2^2$ & $\gamma \|y_k\|_{1} + \frac{\delta}{2} \|y_k\|_2^2$ &  $0< \mu < \frac{1}{1 + {N_{\max}}/{\delta}}$ \\ 

\hline 
 \rule[0ex]{0pt}{4ex} $\frac{1}{2}\|u\|_2^2$ & $\gamma \|y_k\|_{1,+} + \frac{\delta}{2} \|y_k\|_2^2$ &  $0< \mu < \frac{1}{1 + {N_{\max}}/{\delta}}$  \\ 
 
\hline
Huber loss  & $\gamma \|y_k\|_{1,+} + \frac{\delta}{2} \|y_k\|_2^2$ &  $0< \mu < \frac{1}{\eta + {N_{\max}}/{\delta}}$ \\ 

\hline \hline
\end{tabular} 

\begin{tablenotes}

		\vspace{0.5em}
	\item[a]
		$N_{\max}  \defeq \max_{1 \le k \le N} N_k$ is the maximum number of the dictionary atoms 
		that are allowed at each agent.
	
\end{tablenotes}

\end{threeparttable}

\end{table}

\subsection{Novel Document Detection via Dictionary Learning}
\label{Sec:Experiment:NovelDocumentDetection}

In the novel document detection application \cite{Kasiviswanathan12,Aiello2013sensing,Takahashi2014Jan}, a stream of documents arrives in blocks at the network, and the task is to detect which of the documents in the incoming batch are associated with topics that have not been observed previously, and to incorporate the new block of data into the knowledge database to detect new topics/documents in future incoming batches. We refer to each such step as a ``time-step'' and we use $x_t^s$ to denote the $t$th data sample in the $s$th time-step, where $1 \le t \le T_s$ with $T_s$ being the number of samples in the $s$th time-step ($T_s=1000$ for all $s$ in this example), and $1\leq s \leq 8$ since our dataset only contains enough data for eight time-steps. We simulate our dictionary learning algorithm using the Huber cost function as the residual metric. We compare our algorithm performance to that proposed in \cite{Kasiviswanathan12} under the same setup proposed there. The data is from the TDT2 dataset, which contains news documents associated with their dominant topics collected over the first $27$ weeks of $1998$. The data is compiled into a term frequency-inverse document frequency (TF-IDF) matrix $X \in \mathbb{R}^{M \times T}$, where $M = 19527$ and $T = 9394$. The documents have been processed so that only the most frequent $30$ topics (and documents associated with them) are preserved. In this experiment, we allow all agents in the network to observe the incoming data. The key observation is that if a document belongs to a topic that has been observed previously, then it is expected that the objective value of the optimization problem \eqref{Equ:DictLearnDist:NewInference_Objective}--\eqref{Equ:DictLearnDist:NewInference_Constraint} will be ``small'' since the document should be well modeled by the available dictionary. On the other hand, when the objective value is ``large,'' then this is an indication that the document is not well modeled by the available dictionary.


	In this application, we let $f(u) = \sum_{m=1}^M L(u_m)$, where $L(u_m)$ is chosen to be the scalar Huber function defined in Table \ref{Tab:Task}. We choose Huber loss for the following reasons. {The work \cite{Kasiviswanathan12} points out that some of the coefficients of the representation error $u = x_t-W y$ in text documents contain large, impulsive values. For this reason,  the work \cite{Kasiviswanathan12} adopts the $\ell_1$ loss $f(u) = \|u\|_1$ because this loss grows only linearly for large $u$ and is less sensitive to large outliers. However, $\ell_1$ loss is not differentiable and has a conjugate function of  zero with domain $\mc{V}_f = \{ \nu: \|\nu\|_{\infty} \le 1\}$. In comparison, the Huber loss, while preserving the linear growth for large $u$, is smooth and has Lipschitz gradients, which gives a quadratic conjugate function (see Tab. \ref{Tab:ConjProx} and Sec. \ref{Sec:DictLearn:ChoiceResReg}) that naturally regularizes the dual cost \eqref{Equ:DictLearn:Inference_Dual_Objective} to make it strongly convex. In this way, we end up with a better conditioned optimization problem, which allows first-order methods (e.g., diffusion) to achieve relatively fast convergence and satisfactory performance on the dual inference problem \eqref{Equ:DictLearn:Inference_Dual_Objective}--\eqref{Equ:DictLearn:Inference_Dual_Constraint}.
} The setup is the same as in \cite{Kasiviswanathan12},\footnote{We would like to thank S. P. Kasiviswanathan for kindly sharing his MATLAB code through e-mail communication in order to reproduce the simulation in \cite{Kasiviswanathan12}, including the ordered data.} except that we start with only ten dictionary atoms, and add ten additional atoms after each time-step. We simulate the last line of the non-negative matrix factorization setup in Table~\ref{Tab:Task}. We compare our algorithm to the one proposed in \cite{Kasiviswanathan12}, which simulates the setup where $f(u) = \|u\|_1$, $h_y(y)=\|y\|_1$, and $\mathcal{W}_k = \{w: \|w\|_1 \leq 1\}$. { Therefore, the choice of the penalty function $f(u)$ is also slightly different, as we use Huber loss while \cite{Kasiviswanathan12} uses $\ell_1$ loss.}

	For the simulation of the diffusion algorithm, the data are normalized so that $\|x_t^s\|_2 = 1$. In contrast, when testing on the centralized ADMM-based algorithm from \cite{Kasiviswanathan12}, the data are normalized so that $\|x_t^s\|_1 = 1$ in keeping with the proposed simulation setup there. The constraint set for $W$ for the diffusion-based algorithm is $\left\{W: \| [ W ]_{:,q} \|_2 \le 1, \; W \succeq 0\right\}$, while the constraint set for the ADMM-based algorithm from \cite{Kasiviswanathan12} is $\left\{W: \| [ W ]_{:,q} \|_1 \le 1, \; W \succeq 0\right\}$. We choose $\gamma = 0.05$ and $\delta = 0.1$. For the initialization of the dictionary for the ADMM  algorithm from \cite{Kasiviswanathan12}, we let the algorithm iterate between the sparse coding step and the dictionary learning step $35$ times. The diffusion algorithm runs through the data once. We choose $\eta = 0.2$ for the connection point between the quadratic part and the linear part of the Huber loss function. Both the fully connected and distributed algorithms utilize a learning step-size of $\mu_w(s) = 1/s$, where $s$ is the current time-step for learning of the dictionary. For the inference, the fully connected algorithm utilizes $\mu^\textrm{FC} = 0.5$, while the distributed algorithm uses $\mu = 0.05$. The fully connected algorithm performs $100$ iterations for the inference, while the distributed algorithm utilizes $1000$ iterations for the inference. Samples $1$-$1000$ are used for the initialization of the dictionary. Novel documents are only introduced at the first (samples $1001$-$2000$), second ($2001$-$3000$), fifth ($5001$-$6000$), sixth ($6001$-$7000$), and eighth ($8001$-$9000$) time-steps. For this reason, we only execute the novel document detection part of the algorithm at those time-steps, and present the ROC curves for those time-steps. We run our algorithm using the fully connected case, where $A = \frac{1}{N} \mathds{1} \mathds{1}^T$ and the distributed case where the probability that two nodes are connected is $0.5$, and the combination matrix is the Metropolis rule. 
	
To obtain the distributed algorithm, we note from \eqref{Equ:DictLearn:Split_Nodes} that
	\begin{align}
		J_k(\nu; x_t^s) \defeq	
								\displaystyle
								\frac{1}{N} (f^{\star}(\nu) - \nu^T x_t^s)
								\!+\! 
								h_{y_k}^{\star}(w_k^T \nu)
							\label{Equ:DictLearn:Experiments:NovelDocument:HuberResidual:J_k}
	\end{align}
where we are using $w_k$ instead of $W_k$ because each agent $k$ is in charge of one atom of the dictionary (i.e., the $k$-th column of $W$).
Since we now use $f(u) = \sum_{m=1}^M L(u_m)$ and $h_{y_k}(y_k) = \gamma \|y\|_{1,+} + \frac{\delta}{2} \|y\|_2^2$ (according to the last row of Table~\ref{Tab:Task}), we obtain that $f^\star(\nu) = \frac{\eta}{2} \|\nu\|_2^2$, $\mathcal{V}_f = \{\nu: \|\nu\|_\infty \leq 1\}$, and $h_{y_k}^{\star}(w_k^T \nu) = \mathcal{S}_{\frac{\gamma}{\delta}}^+\left(\frac{w_k^T \nu}{\delta}\right)$ according to Table~\ref{Tab:ConjProx}. A straightforward calculation then shows that 
	\begin{align}
		\nabla_\nu f^\star(\nu) &= \eta\cdot \nu,
		\quad
		\nabla_\nu h_{y_k}^{\star}(w_k^T \nu) =  \frac{1}{\delta} \mathcal{T}_\gamma^+ (w_k^T \nu) w_k \label{Equ:DictLearn:Experiments:NovelDocument:HuberResidual:nabla_h_star}
	\end{align}
	Substituting \eqref{Equ:DictLearn:Experiments:NovelDocument:HuberResidual:nabla_h_star} into the gradient of \eqref{Equ:DictLearn:Experiments:NovelDocument:HuberResidual:J_k}, we obtain:
	\begin{align}
		\nabla_\nu J_k(\nu;x_t) &= \frac{1}{N} (\eta\cdot \nu - x_t)
								\!+\! 
								\frac{1}{\delta} \mathcal{T}_\gamma^+(w_k^T \nu) w_k \label{Equ:DictLearn:Experiments:NovelDocument:HuberResidual:nabla_J_k}
	\end{align}	
	where we let $\mathcal{N}_I = \mathcal{N}$ and all agents in the network have access to $x_t^s$. By substituting \eqref{Equ:DictLearn:Experiments:NovelDocument:HuberResidual:nabla_J_k} into the inference part of Alg.~\ref{alg:DiffusionSparseCoding}, we immediately obtain the inference part of Alg.~\ref{alg:DocumentDetection:Huber}. For the learning portion of the algorithm, we need to compute $y_{k,t}^o$ at node $k$ once $\nu_{t}^o$ has been estimated. With our choices of $f(u)$ and $h(y_k)$, we observe from Table \ref{Tab:ConjProx} that $y_{k,t}^o$ may be obtained as $y^o_{k,t} = \mathcal{T}_{\frac{\gamma}{\delta}}^+\left(\frac{w_k^T \nu_t^o}{\delta}\right) = \frac{1}{\delta} \mathcal{T}_{\gamma}^+\left(w_k^T \nu_t^o\right)$ (as listed in Alg.~\ref{alg:DocumentDetection:Huber}). Now, using the fact that $h_{w_k}(w_k) = 0$ (see Table~\ref{Tab:Task}), we have that the update rule for $w_k$ from Alg.~\ref{alg:DiffusionSparseCoding} becomes
	\begin{align}
		w_{k,t} = \Pi_{\mathcal{W}_k}\left\{w_{k,t-1} \!+\! \mu_w \nu_t^o y_{k,t}^{o}\right\}
		\label{Equ:DictLearn:Experiments:NovelDocument:DictUpdate}
	\end{align}
	where $\mathcal{W}_k = \{w: \|w\|_2 \leq 1, w \succeq 0\}$ (see Table~\ref{Tab:Task}). 
{ When recursion \eqref{Equ:DictLearn:Experiments:NovelDocument:DictUpdate} finishes going through  the data samples in the $s$-th time-step, the most up-to-date dictionary is denoted by 
$W^s = [w_1^s \cdots w_N^s]$.}

In this example, we do not need to recover $z_t^o$ in \eqref{Equ:DictLearnDist:z_optimal_primal}, but we only need to recover the cost value for representing a test data sample $\xi_t$ using dictionary $W^s$ learned up to the $s$-th time-step:
	\begin{align}
		\min_{\{y_k\}} 
		\left[
			f\Big(
				\xi_t - \sum_{k=1}^N w_{k}^s y_k
			\Big)
			+
			\sum_{k=1}^N 
			h_{y_k}(y_k)
		\right]
		\label{Equ:DictLearn:Experiment:NovelDocument:PrimalCostValue}
	\end{align}
where we use $\xi_t$ to differentiate it from the training data sample $x_t^s$.
Interestingly, since strong duality holds for this example, based on the argument from \eqref{Equ:DictLearnDist:NewInference_Objective} to \eqref{Equ:DictLearn:Inference_Dual_Objective}, the above minimum primal cost \eqref{Equ:DictLearn:Experiment:NovelDocument:PrimalCostValue} is equal to the maximum value of its associated dual cost:
	\begin{align}
		\max_{\nu} g(\nu, \xi_t) = g(\nu_t^o,\xi_t) = -\sum_{k=1}^N J_k(\nu_t^o,\xi_t)
		\label{Equ:DictLearn:Experiment:NovelDocument:DualCostValue}
	\end{align}
where the first equality follows from the fact that $\nu_t^o$ is the optimizer of the dual problem.
Therefore, we can obtain the minimum primal cost \eqref{Equ:DictLearn:Experiment:NovelDocument:PrimalCostValue} by computing the maximum dual cost \eqref{Equ:DictLearn:Experiment:NovelDocument:DualCostValue}, which can be done in many ways with one of them being the diffusion strategy. In order to obtain a scaled multiple of \eqref{Equ:DictLearn:Experiment:NovelDocument:DualCostValue}, we setup the following scalar optimization problem:
\begin{align}
	\min_g\   \sum_{k=1}^N V_k(g)
	\label{Equ:DictLearn:Experiments:NovelDocument:SquareResidual:consensus}
\end{align}
where 
	\begin{align}
		V_k(g) \triangleq \frac{1}{2} \left(J_k(\nu_t^o,\xi_t)+g\right)^2
	\end{align}
from which we can obtain the following scalar diffusion algorithm \cite{chen2013JSTSPpareto}:
\begin{align}
\begin{cases}
	\phi_{k}(i) = g_{k}(i-1) - \mu_g (J_k(\nu_t^o,\xi_t)+g_{k}(i-1))\\
	\displaystyle g_{k}(i) = \sum_{\ell \in \mathcal{N}_k} a_{\ell k} \phi_{\ell}(i)
	\end{cases}
	\label{Equ:DictLearn:Experiments:NovelDocument:SquareResidual:ScalarDiffusion}
\end{align}
After sufficient iterations, recursion \eqref{Equ:DictLearn:Experiments:NovelDocument:SquareResidual:ScalarDiffusion} approximates the minimizer of \eqref{Equ:DictLearn:Experiments:NovelDocument:SquareResidual:consensus}, which is $g^o_t = -\frac{1}{N} \sum_{k=1}^N J_k(\nu_t^o,\xi_t)$. { Comparing $g^o_t$ to \eqref{Equ:DictLearn:Experiment:NovelDocument:DualCostValue}, we note that there is an additional positive scaling factor, $1/N$, in $g^o_t $. However, it does not affect the result since it can be absorbed into the threshold parameter:
	\begin{align}		
		-\sum_{k=1}^N J_k(\nu_t^o,\xi_t)
		\underset{H_0}{\overset{H_1}{\gtrless}} 	\chi'
		\quad\Leftrightarrow\quad		
		&g^o_t	\underset{H_0}{\overset{H_1}{\gtrless}} 	\chi \defeq \frac{\chi'}{N}
	\end{align}
where $H_1$ and $H_0$ denote the hypotheses of ``the document is novel'' and ``the document is not novel'', respectively. In other words, using a threshold $\chi'$ for the original cost \eqref{Equ:DictLearn:Experiment:NovelDocument:DualCostValue}, is equivalent to using the threshold $\chi = \chi'/N$ for $g_t^o$.
 	
}

	The final algorithm is listed in Alg.~\ref{alg:DocumentDetection:Huber}. Each node in the network is responsible for a single dictionary atom. The sparse coding stages of the centralized ADMM-based algorithm from \cite{Kasiviswanathan12} utilize $35$ iterations, and the number of iterations of the dictionary update steps are capped at $10$ for all iterations other than the initialization step, which are the default setup in the code of \cite{Kasiviswanathan12}. We observe that the performance of the centralized ADMM-based algorithm reproduced in this manuscript is competitive with that in \cite{Kasiviswanathan12}, even though the initial dictionary size is chosen to be ten, as opposed to $200$ atoms (as was done in the experiment in \cite{Kasiviswanathan12}). Furthermore, for our algorithm, since we are simulating a network of $N$-agents on a single machine, we expect the computation time to be $N$ times as much as that in \cite{Kasiviswanathan12} in order to have a fair comparison. This is because the gradient descent steps and the combination steps in \eqref{Equ:DictLearnDist:ATC_adapt}--\eqref{Equ:DictLearnDist:ATC_combine} should be finished concurrently in an actual $N$-agent network, while our single-machine simulation can only perform them sequentially. For this reason, we choose the setup for our algorithm (such as the number of inference iterations) to be about $N$ times of that in \cite{Kasiviswanathan12} to ensure a fair comparison.\footnote{When applying the centralized gradient descent to the dual inference problem \eqref{Equ:DictLearnDist:DualProblem_Objective_newForm}--\eqref{Equ:DictLearnDist:DualProblem_Constraint_newForm} with $1000$ iterations at a single machine, we found that the entire learning time over one time-step ($1000$ samples) is approximately the same as that of the ADMM-based method from \cite{Kasiviswanathan12} using the same MATLAB implementation for the time benchmark.}


\noindent\begin{algorithm}[t]
	\caption{{\small{Model-distributed diffusion strategy for distributed novel document detection (Huber Loss Residual).}}}
	\label{alg:DocumentDetection:Huber}
	{\small{
	\begin{algorithmic}
	{
	\STATE {\bf Initialization:} The sub-dictionaries $\{W_k\}$ are randomly initialized and then projected onto \eqref{Equ:ProbForm:W_subUnitNormNonnegConstraint} using \eqref{Equ:DictLearn:DictUpdate_Projection_nonnegativenormball}.}
	
	\FOR{each time step $s = 1, 2, \ldots, 8$}
	\STATE \underline{\textbf{\emph{Dictionary Learning:}}}
	\FOR{each training sample $x_t^s$ from time-step $s$, ($t=1,\ldots, T_s$)}
		\STATE Each node $k$ repeats until convergence:
			\!\!\!\!{\footnotesize{\begin{equation*}
			\begin{cases}
				\!\psi_{k,i} \!\!=\! \nu_{k,i-1} \!-\!\!  \frac{\mu}{N} (\!\eta \nu_{k,i-1} \!-\! x_t^s\!) \!-\! \frac{\mu}{\delta} \mathcal{T}_\gamma^+(\!w_{k,t-1}^T \nu_{k,i-1}\!)w_{k,t-1}\\
				\!\nu_{k,i} \!=\! \Pi_{\nu \in [-1,1]}\left\{\sum_{\ell \in \mathcal{N}_k} a_{\ell k} \psi_{\ell,i}\right\}
			\end{cases}
			\end{equation*}}}%
		{with initialization $\{\nu_{k,0} = 0, \; k=1,\ldots,N\}$.}
		where the above projection is carried out according to \eqref{Equ:DictLearnDist:Projection_Inf_Norm}.
		\STATE Set $\nu_t^o = \nu_{k,i}$. Compute $y_{k,t}^o = \frac{1}{\delta}\mathcal{T}_\gamma^+(w_{k,t-1}^T \nu_t^o)$.
		\STATE Update the dictionary using:
		\begin{equation*}
			w_{k,t} = \Pi_{\|w\|_2 \leq 1} \left\{\Pi_{w \succeq 0}\left\{w_{k,t-1} \!+\! \mu_w(s) \nu_t^o y_{k,t}^{o }\right\}\right\}
		\end{equation*}
	\ENDFOR
	\STATE Let $w_{k}^s$ denote the most up-to-date sub-dictionary at agent $k$.
	\STATE \underline{\textbf{\emph{Novel Document Detection:}}}
	\FOR{each test data sample $\xi_t$, each node $k$}
		\STATE Repeat until convergence:
			\!\!\!\!{\footnotesize{\begin{equation*}
			\begin{cases}
				\!\psi_{k,i} \!\!=\! \nu_{k,i-1} \!\!-\!\!  \frac{\mu}{N} (\!\eta \nu_{k,i-1} \!-\! \xi_t\!) \!-\! \frac{\mu}{\delta} \mathcal{T}_\gamma^+\big(\!(w_{k}^s)^T \nu_{k,i-1}\!\big)w_{k}^s\\
				\!\nu_{k,i} = \Pi_{\nu \in [-1,1]}\left\{\sum_{\ell \in \mathcal{N}_k} a_{\ell k} \psi_{\ell,i}\right\}
			\end{cases}
			\end{equation*}}}
		\STATE Set $\nu_t^o = \nu_{k,i}$. 
		\STATE Perform diffusion strategy to optimize \eqref{Equ:DictLearn:Experiments:NovelDocument:SquareResidual:consensus} until convergence:
		\begin{align*}
			\begin{cases}
	\phi_{k}(i) = g_{k}(i-1) - \mu_g (J_k(\nu_t^o,\xi_t)+g_{k}(i-1))\\
	g_{k}(i) = \sum_{\ell \in \mathcal{N}_k} a_{\ell k} \phi_{\ell}(i)
	\end{cases}
		\end{align*}
		where $J_k(\nu,\cdot)$ is defined in \eqref{Equ:DictLearn:Experiments:NovelDocument:HuberResidual:J_k}.		
		\STATE Set $g^o_t = g_{k,i}$.
		\IF{$g^o_t > \chi$}
			\STATE declare document as novel.
		\ELSE
			\STATE declare document as not novel.
		\ENDIF
	\ENDFOR
	\STATE Add nodes to network (expand the dictionary)
	\ENDFOR
	\end{algorithmic}}}
\end{algorithm}

\begin{figure}[th!]
\centering
	\includegraphics[width=0.5\textwidth]{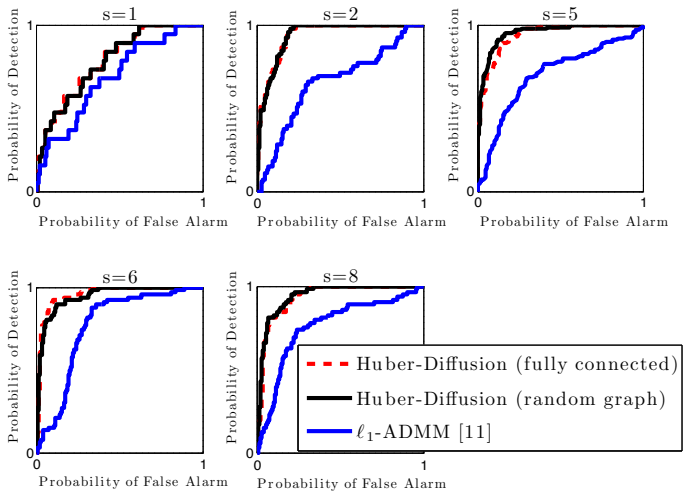}
	\caption{Application of dictionary learning to novel document/topic detection. At each time step, the algorithms receive $1000$ documents. The task is to determine which documents are associated with topics that have already been observed, and which documents are associated with topics that have not yet been observed. These curves represent the ROC curve associated with each time step against a changing test set. The $x$-axis represents probability of false alarm while the $y$-axis represents probability of detection. The area under each cuve is listed in Table~\ref{Tab:DocumentDetection:L1:AUC}.}
	\label{Fig:DocumentDetection:L1}
\end{figure}

\begin{table}[t!]
	\caption{Area under ROC curve for the three tested algorithms. Novel documents not presented in time-steps $3$, $5$, $7$.}
	\label{Tab:DocumentDetection:L1:AUC}
	\begin{center}
	\begin{tabular}{c||c|c|c}
	\hline \hline
	\rowcolor[gray]{0.9} Time Step & ADMM \cite{Kasiviswanathan12} & Diffusion (Fully Connected) & Diffusion \\ 
	\hline 
	$1$ & $0.69$ & $\textbf{0.79}$ & $\textbf{0.79}$ \\ 
	\hline 
	$2$ & $0.65$ & $\textbf{0.94}$ & $0.93$ \\ 
	\hline 
	$5$ & $0.70$ & $0.94$ & $\textbf{0.95}$ \\ 
	\hline 
	$6$ & $0.77$ & $\textbf{0.96}$ & $0.95$ \\ 
	\hline 
	$8$ & $0.76$ & $0.93$ & $\textbf{0.94}$ \\ 
	\hline \hline
	\end{tabular} 
	\end{center}\vspace{-1\baselineskip}
\end{table}

The performance of the algorithms is illustrated in Fig.~\ref{Fig:DocumentDetection:L1}. We observe that the Huber loss function improves performance relative to the $\ell_1$ function. The area under each ROC curve is listed in Table~\ref{Tab:DocumentDetection:L1:AUC}. Since the different algorithms were initialized with different dictionaries, it may be possible for the sparsely-connected diffusion strategy to slightly outperform the fully-connected diffusion strategy. We observe this effect in Table~\ref{Tab:DocumentDetection:L1:AUC}, where the sparsely-connected network outperforms the fully-connected network by $0.01$ (area under ROC curve).

\subsection{Biclustering via Sparse Singular-Value-Decomposition}
\label{Sec:Experiment:BiClustering}
Consider next the cancer data matrix $X \in \mathbb{R}^{M\times T}$ from \cite{lee2010biclustering}, where $M=56$ and $T=12,625$. Each row of $X$ contains the genetic information for each of $56$ patients. Each patient belongs to one of four cancer categories: Normal,  Carcinoid, Colon, and SmallCell. The algorithm is unaware of the true category (label) of any patient, but wants to cluster patients into groups with different cancer types using the genetic information. The problem was formulated in \cite{lee2013distributed} as a bi-clustering task (see also Tables \ref{Tab:Task}--\ref{Tab:ConjProx}) that factorizes $X$ as
\begin{align}
	X 	\approx 	\sum_{k=1}^N w_{k} y_{k}^T 
	\label{eq:SVD}
\end{align}
with both $w_k \in \mathbb{R}^{M \times 1}$ and $y_k \in \mathbb{R}^{T \times 1}$ being sparse. 

\noindent\begin{algorithm}
	\caption{\small Simplified algorithm from \cite{lee2010biclustering} for biclustering.}
	\label{alg:lee}
	\begin{algorithmic}
	{\small
	\FOR{each $k$}
	\STATE Apply standard SVD to $X = w_\textrm{old} s_\textrm{old} y_\textrm{old}^T$. Repeat until convergence:
	\begin{enumerate}
		\STATE Set $\tilde{y} = \mathcal{T}_{\lambda}(X^T w_\textrm{old})$, 
			and $y_\textrm{new} = \tilde{y}/\|\tilde{y}\|_2$.
		\STATE Set $\tilde{w} = \mathcal{T}_{\beta}(X y_\textrm{new})$,
			and $w_\textrm{new} = \tilde{w}/\|\tilde{y}\|_2$.
		\STATE Set $w_\textrm{old} = w_\textrm{new}$.
	\end{enumerate}
	\STATE Set $w_k = w_\textrm{new}$, $s_k=w_\textrm{new}^T X y_\textrm{new}$, and $y_k = s_k y_\textrm{new}$.
	\STATE Set $X = X - w_k y_k^T$.
	\ENDFOR
	}
	\end{algorithmic}
\end{algorithm}

\noindent\begin{algorithm}
	\caption{\small Model-distributed diffusion strategy for { online} biclustering.}
	\label{alg:biclustering}
	\begin{algorithmic}
	{\small
	
	\STATE {\bf Initialization:} The sub-dictionaries $\{W_k\}$ are randomly initialized and then projected onto \eqref{Equ:ProbForm:W_subUnitNormConstraint} using \eqref{Equ:DictLearn:DictUpdate_Projection_normball}.

	\FOR{each input data sample $x_t$, each node $k$}
		\STATE Repeat until convergence:
			\!\!\!\!\begin{equation*}
			\begin{cases}
				\!\psi_{k,i} \!\!=\! \nu_{k,i-1} \!-\! \mu_\nu \frac{1}{N} (\nu_{k,i-1} - x_t) -\\
								\!\!\quad\quad\quad\quad\quad\quad  \frac{\mu_\nu}{\delta} \mathcal{T}_\gamma(w_{k,t-1}^T \nu_{k,i-1})w_{k,t-1}\\
				\!\nu_{k,i} = \sum_{\ell \in \mathcal{N}_k} a_{\ell k} \psi_{\ell,i}
			\end{cases}
			\end{equation*}
		{with initialization $\{\nu_{k,0}=0, \;k=1,\ldots,N\}$.}
		\STATE Set $\nu_k^o = \nu_{k,i}$. Compute $y_k^o = \frac{1}{\delta}\mathcal{T}_\gamma(w_{k,t-1}^T \nu^o)$.
		\STATE Update the dictionary using:
		\begin{equation*}
			w_{k,t} = \Pi_{\|w\| \leq 1}\left\{\mathcal{T}_\beta\left(w_{k,t-1} \!+\! \mu_w \nu_k^o y_k^{o T}\right)\right\}
		\end{equation*}
	\ENDFOR
	}
	\end{algorithmic}
\end{algorithm}

\begin{figure}[th!]
	\centerline{
		\subfigure[Data clusters obtained by Alg.~\ref{alg:lee}.]{
	\includegraphics[width=0.45\textwidth]{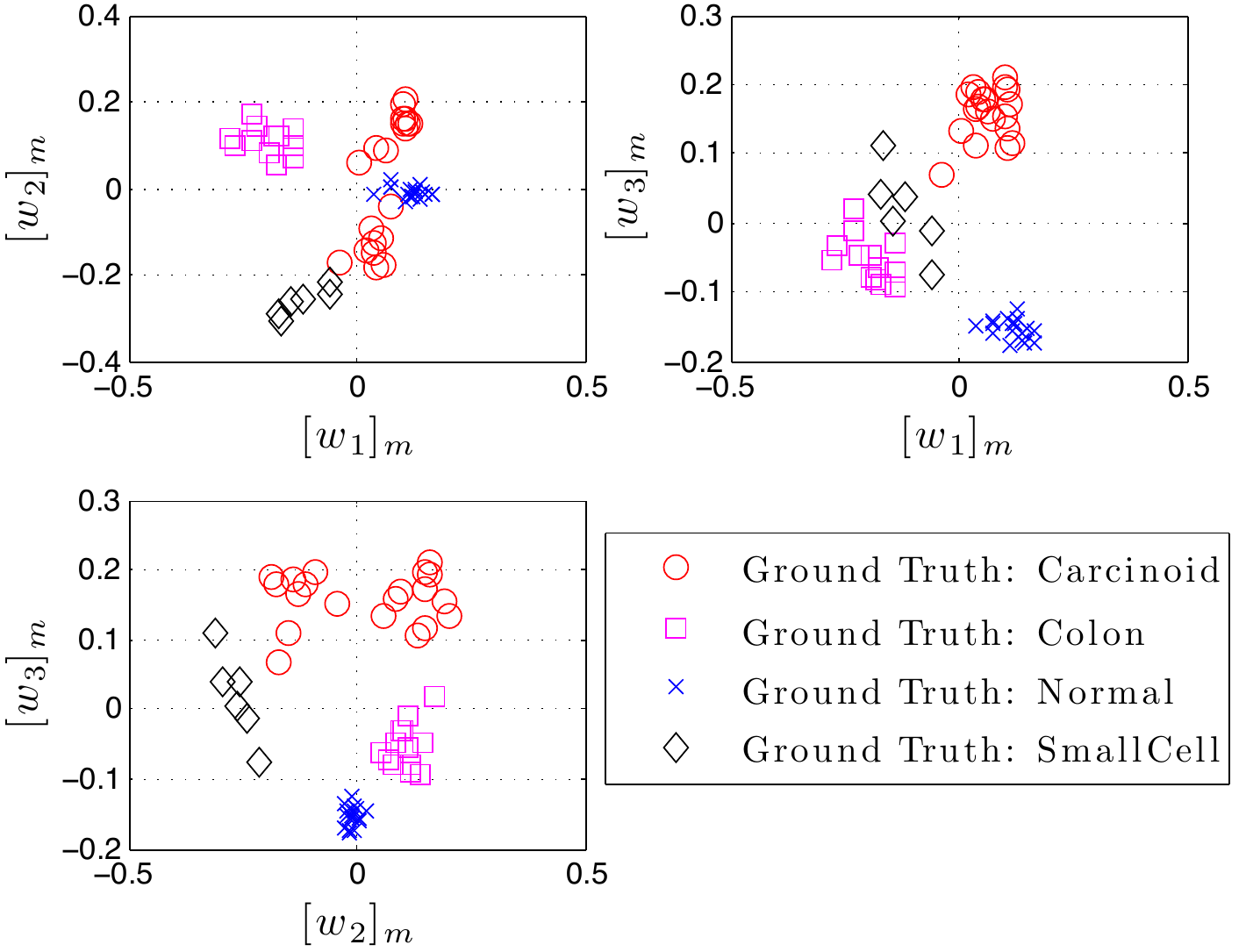}
		}
	}\vspace{0.5em}
	\centerline{
		\subfigure[Data clusters obtained by Alg.~\ref{alg:biclustering}.]{
			\includegraphics[width=0.45\textwidth]{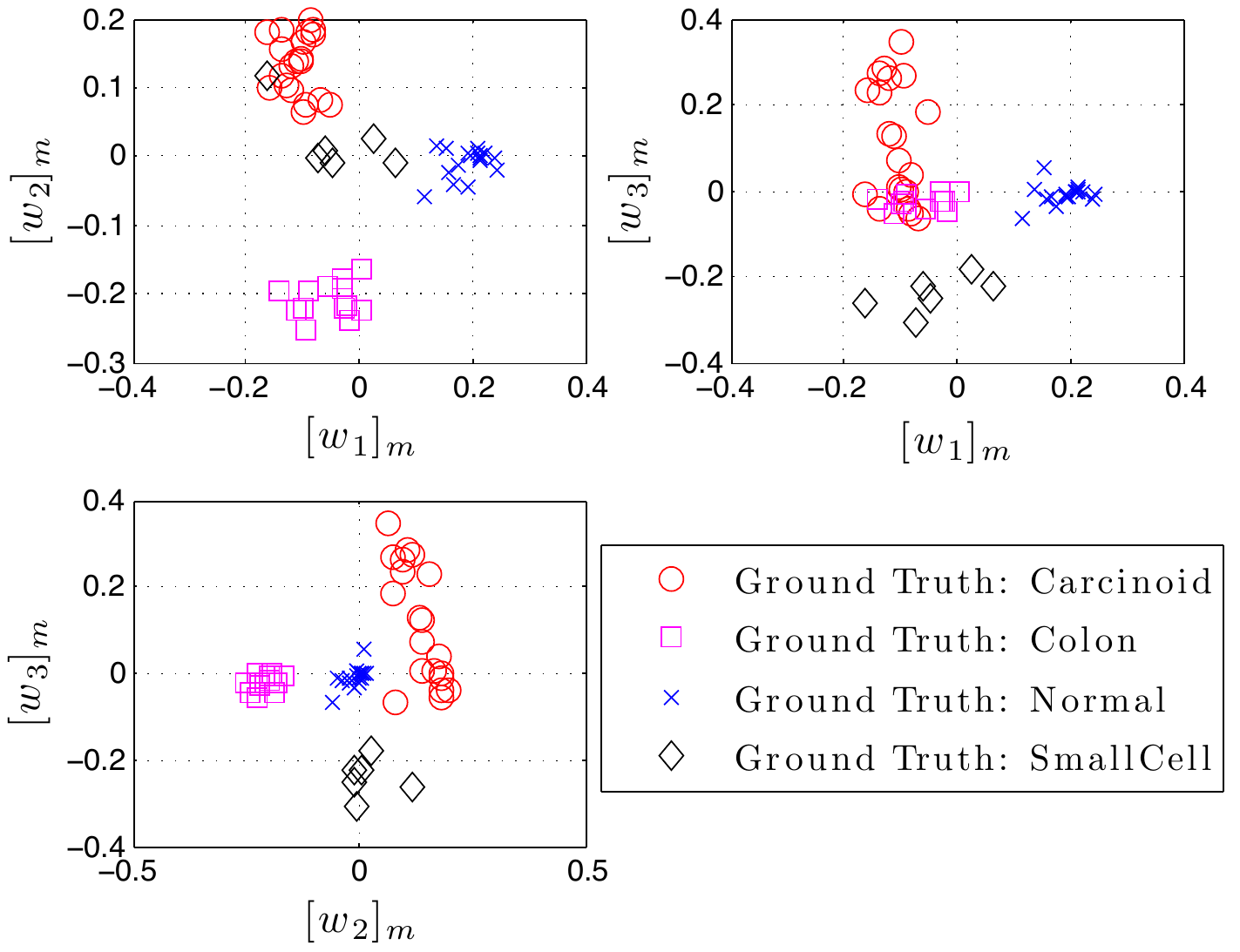}
		}
	}
	\caption{Application of microarray biclustering. Each marker represents one patient, and $[w_k]_m$ denotes the $m$-th entry of the dictionary atom $w_k$, where $m=1,\ldots, 56$ is the index of the patients and $k=1,2,3$ is the index of the dictionary atoms. The algorithm is unaware of the ground truth of the cancer categories of each patient. { After the bi-clustering is done, we add colors to different markers according to the ground truth (label) to visualize the success of the bi-clustering task.}}
	\label{Fig:biclustering_clusters}
\end{figure}

In Alg.~\ref{alg:lee}, we list the algorithm from \cite{lee2010biclustering}, which alternates between two sparse coding steps to obtain $y$ and $w$, respectively. Observe that the algorithm is a batch algorithm, in that it utilizes the entire data set at each iteration. In addition, the algorithm works by computing the best sparse rank-$1$ approximation for the matrix $X \approx w_1 y_1^T$, then computes the best rank-$1$ approximation for $X-w_1 y_1^T \approx w_2 y_2^T$, and so on.  In contrast, our proposed Alg. \ref{alg:DiffusionSparseCoding}, when specialized to the bi-clustering application (see Alg.~\ref{alg:biclustering}), runs through the data in an online manner and obtains the $\{w_k\}$ simultaneously.

We choose $N=3$ to be consistent with the setup in \cite{lee2010biclustering}, where each node is responsible for a single dictionary atom. We set $\gamma = 0.5$ and $\beta = 0.01$. Since the number of nodes is small, we simulate the fully connected case where the combination matrix $A = \frac{1}{N} \mathds{1}_N \mathds{1}_N^T$ (i.e., each node is effectively averaging the estimate of $\psi_{k,i}$). We run Alg.~\ref{alg:lee} until $\|w_\textrm{new}-w_\textrm{old}\|_\infty < 1\times 10^{-10}$. We run our algorithm's sparse coding for a total of $2000$ iterations. We choose $\mu_\nu = 0.01$, $\mu_w = 5\times 10^{-3}$, and $\delta = 0.01$. In Fig.~\ref{Fig:biclustering_clusters}, we plot, in the same manner as\cite{lee2010biclustering}, the clustering results of Algorithms \ref{alg:lee}--\ref{alg:biclustering}. This is an unsupervised learning task, meaning that, during the learning process, the algorithms are unaware of the ground truth of the cancer categories of each patient. Still, the algorithms are required to cluster patients into different groups according to their underlying genetic information, hoping that patients of similar genetic information will be clustered together. After the clustering is done, we add colors to different markers according to the ground truth (label) to visualize and evaluate the result of the clustering. The clustering will be more successful if (i) markers of the same color are clustered together, and (ii) markers of different colors are well separated. We observe that both algorithms, without the use of the cancer labels, can successfully cluster the data into 4 distinct clusters according to the genetic information, with each cluster corresponding to a different type of cancer.  The advantage of the diffusion strategy is that it only requires each node to observe each data sample (each column of $X$) once (not batched) and obtain $\{w_1,w_2,w_3\}$ simultaneously. Note that in this example an additional data collection process is required to gather all the $w_1$, $w_2$, and $w_3$ to generate the final bi-clustering plots in Fig. \ref{Fig:biclustering_clusters}. This is because we need to use $([w_1]_m, [w_2]_m, [w_3]_m)$ to represent the genetic profile of each patient $m$. This step is usually less demanding than learning the $\{w_k\}$, especially in large-scale genetic data analysis. The agents may choose to report the obtained results periodically. Nevertheless, the computation-intensive learning process in bi-clustering is still distributed over the network, where the agents learn different $\{w_k\}$ in an online and simultaneous manner.

\section{Conclusion}
\label{Sec:Conclusion}

In this paper, we studied the online dictionary learning problem over distributed models, where each agent is in charge of a portion of the dictionary atoms and the agents collaborate to represent the data. Using the concepts of conjugate function and dual decomposition, we transform the original learning problem into a form that is amenable to distributed optimization, which is then solved by means of a diffusion strategy. The collaborative inference step generates dual variables that are used by the agents to update their dictionary atoms without the need to share their dictionaries or even the coefficient models for the training data. The proposed algorithm is tested over two typical tasks of dictionary learning, namely, novel document detection and bi-clustering. The results demonstrate that our proposed algorithm can solve the dictionary learning tasks effectively in a distributed and online manner.

In relation to the convergence behavior, we remark that the general learning problem \eqref{Equ:ProbForm:DictLearn_Objective}--\eqref{Equ:ProbForm:DictLearn_Constraint} is not jointly convex with respect to both $W$ and $y$. This fact explains why convergence guarantees  towards a global minimum, when it exists, are generally not available in the literature. A common technique for solving such coupled optimization problems is to alternate between the minimization over one variable  while keeping the other variable fixed. In this article, we followed a similar construction albeit one that operates in an online and distributed manner. For the inference problem \eqref{Equ:ProbForm:InferenceProblem}, we applied the diffusion strategy, which has already been shown in prior studies \cite{chen2013JSTSPpareto} to converge within $O(\mu^2)$ to the optimal inference solution. For the dictionary update step, we used a proximal projection step.  Simulation results in this article and by other authors have indicated that such alternating optimization solutions tend to perform well in practice.

\appendices

\section{Derivation of Some Typical Conjugate Functions}
\label{Appendix:DerivationConjugateFun}

In this appendix, we derive the conjugate functions listed in Table \ref{Tab:ConjProx}. The conjugate functions for $\frac{1}{2}\| u \|_2^2$, and their corresponding domains can be found in \cite[pp.90-94]{boyd2004convex}. The conjugate function for the scalar Huber loss $L(u_m)$ can be found in \cite{zach2010practical} as $L^{\star}(\nu_m)	=	\frac{1}{2} \nu_m^2$ with $| \nu_m | \le 1$.
Therefore, by the ``sums of independent functions'' property\footnote{\label{FN:ConjProperty_SumOfIndependentFunc}If $f(x_1,\ldots,x_N)=f_1(x_1) + \cdots f_N(x_N)$, then the conjugate function for $f(x_1,\ldots,x_N)$ is given by $f^{\star}(\nu_1,\ldots,\nu_N) = f_1^{\star}(\nu_1) + \cdots + f_N^{\star}(\nu_N)$, where $f_1^{\star}(\nu_1),\ldots, f_N^{\star}(\nu_N)$ are the conjugate functions for $f_1(x_1), \ldots, f_N(x_N)$, respectively.} in \cite[p.95]{boyd2004convex}, the conjugate function of $\sum_{m=1}^M L(u_m)$ is:
	\begin{align}
		\sum_{m=1}^M L^{\star}(\nu_m)	=	\sum_{m=1}^M \frac{1}{2} \nu_m^2 
									=	\frac{1}{2}\| \nu \|_2^2,
	\end{align}
where the domain is given by
	\begin{align}
		| \nu_m | 	\le 	1, \quad  m =1, \ldots, M 
		\quad \Leftrightarrow \quad
		\| \nu \|_{\infty} \le 1
	\end{align}
	
Next, we derive the conjugate functions for the elastic net regularization term $	h_{y_k}(y_k) = \gamma \| y_k \|_{1} + \frac{\delta}{2} \| y_k \|_2^2$.
By the definition of conjugate functions in \eqref{Equ:DictLearnDist:r_conj_def}, we have
	\begin{align}
		h_{y_k}^{\star}(W_k^T\nu)		&=
									\sup_{y_k}
									\left[
										(W_k^T\nu)^T y_k - h_{y_k}(y_k)
									\right]
									\nn\\
							&=
									-\inf_{y_k}
									\left[
										h_{y_k}(y_k) - (W_k^T\nu)^T y_k 
									\right]
									\nn\\
							&=
									-\inf_{y_k}
									\left[
										\gamma \| y_k \|_{1} 
										\!+\! 
										\frac{\delta}{2} \| y_k \|_2^2 
										\!-\! 
										(W_k^T\nu)^T y_k 
									\right]		
		\label{Equ:Appendix:hyk_conj_interm0}
									\\
							&=
									- \delta 
									\!\cdot\!
									\inf_{y_k}
									\left[
										\frac{\gamma}{\delta} 
										\| y_k \|_{1} 
										\!+\!
										\frac{1}{2} 
										\Big\| y_k \!-\! \frac{1}{\delta} W_k^T\nu \Big\|_2^2 
									\right]
									\nn\\
									&\quad
									+
									\frac{1}{2\delta}
									\| W_k^T \nu \|_2^2
		\label{Equ:Appendix:hyk_conj_interm1}
	\end{align}
where the last step completes the square. Note from \eqref{Equ:DictLearnDist:Prox_def} that the optimal $y_k$ that minimizes the term inside the bracket of \eqref{Equ:Appendix:hyk_conj_interm1} can be expressed as the proximal operator of $(\gamma/\delta)\|y_k\|_1$, which is known to be given by the entry-wise soft-thresholding operator\cite[p.188]{parikh2013proximal} \cite{donoho1995noising}:
	\begin{align}
		y_{k,t}^o
							&=
									\arg\min_{y_k} 
									\left[
										\frac{\gamma}{\delta} 
										\| y_k \|_{1} 
										+ 
										\frac{1}{2} 
										\Big\| y_k \!-\! \frac{1}{\delta} W_k^T\nu \Big\|_2^2 
									\right]
									\nn\\
							&=
									\mathrm{prox}_{\frac{\gamma}{\delta}\|\cdot\|_1}
									\left(
										\frac{W_k^T\nu}{\delta}
									\right)
							=
									\mT_{\frac{\gamma}{\delta}}
									\left(
										\frac{W_k^T\nu}{\delta}
									\right)
		\label{Equ:Appendix:hyk_conj_interm2}
	\end{align}
where $[\mT_{\lambda}(x)]_n 	\defeq 	(| [x]_n |-\lambda)_{+} \mathrm{sgn}([x]_n)$
and $(x)_{+} = \max( x, 0 )$. Substituting \eqref{Equ:Appendix:hyk_conj_interm2} into \eqref{Equ:Appendix:hyk_conj_interm0}, we obtain
	\begin{align}
		h_{y_k}^{\star}(W_k^T\nu)
							&=		
									\mS_{\frac{\gamma}{\delta}}
									\left(
										\frac{W_k^T\nu}{\delta}
									\right)							
	\end{align}
where
	\begin{align}
		\mS_{\frac{\gamma}{\delta}}
		\left(
			x
		\right)	
							&\defeq
									-\gamma
									\!\cdot\!
									\big\|
										\mT_{\frac{\gamma}{\delta}}
										\left(
											x
										\right)
										\!
									\big\|_1
									\!-\!
									\frac{\delta}{2}
									\big\|
										\mT_{\frac{\gamma}{\delta}}
										\left(
											x
										\right)
										\!
									\big\|_2^2
									\!+\!
									\delta
									\cdot 
									x^T
									\mT_{\frac{\gamma}{\delta}}
									\left(
										x
									\right)
	\end{align}
	
Finally, we derive the conjugate function for the nonnegative elastic net regularization function $h_{y_k}(y_k) = \gamma \| y_k \|_{1, + } + \frac{\delta}{2} \|y_k\|_2^2$. Following the same line of argument from \eqref{Equ:Appendix:hyk_conj_interm0}--\eqref{Equ:Appendix:hyk_conj_interm1}, we get
	\begin{subequations}
		\begin{align}
			h_{y_k}^{\star}(W_k^T\nu)		
								&=
										-\inf_{y_k}
										\left[
											\gamma \| y_k \|_{1,+} 
											\!+\! 
											\frac{\delta}{2} \| y_k \|_2^2 
											\!-\! 
											(W_k^T\nu)^T 
											\!
											y_k 
											\!
										\right]		\!\!
			\label{Equ:Appendix:hyk_nonnegative_conj_interm0}
										\\
								&=
										- \delta 
										\!\cdot\!
										\inf_{y_k}
										\left[
											\frac{\gamma}{\delta} 
											\| y_k \|_{1,+} 
											\!+\!
											\frac{1}{2} 
											\Big\| y_k \!-\! \frac{1}{\delta} W_k^T\nu \Big\|_2^2 
										\right]
										\nn\\
										&\quad
										+
										\frac{1}{2\delta}
										\| W_k^T \nu \|_2^2
			\label{Equ:Appendix:hyk_nonnegative_conj_interm1}
		\end{align}
	\end{subequations}
By \eqref{Equ:DictLearnDist:Prox_def}, the optimal $y_{k,t}^o$ that minimizes the term inside the bracket of \eqref{Equ:Appendix:hyk_nonnegative_conj_interm1} is given by 
	\begin{align}
		y_{k,t}^o				&=		\arg\min_{y_k} 
									\left[
										\frac{\gamma}{\delta} 
										\| y_k \|_{1,+} 
										+ 
										\frac{1}{2} 
										\Big\| y_k \!-\! \frac{1}{\delta} W_k^T\nu \Big\|_2^2 
									\right]
		\label{Equ:Appendix:ykto_argmin}
	\end{align}
Applying an argument similar to the one used in \cite{beck2009fast}, 
we can express the optimal $y_{k,t}^o$ in \eqref{Equ:Appendix:ykto_argmin} as
	\begin{align}
		y_{k,t}^o		=		\mT_{\frac{\gamma}{\delta}}^{+}
							\left(
								\frac{W_k^T \nu}{\delta}
							\right)
		\label{Equ:Appendix:ykto_Tplus_expr}
	\end{align}
where $[\mT_{\lambda}^{+}(x)]_n  \defeq 	([x]_n - \lambda)_{+}$.
Substituting \eqref{Equ:Appendix:ykto_Tplus_expr} into \eqref{Equ:Appendix:hyk_nonnegative_conj_interm0}:
	\begin{align}
		h_{y_k}^{\star}(W_k^T\nu)
							&=		
									\mS_{\frac{\gamma}{\delta}}^{+}
									\left(
										\frac{W_k^T\nu}{\delta}
									\right)							
	\end{align}
where
	\begin{align}
		\mS_{\frac{\gamma}{\delta}}^{+}
		\left(
			x
		\right)	
							&\defeq
									-\gamma
									\!\cdot\!
									\big\|
										\mT_{\frac{\gamma}{\delta}}^{+}
										\left(
											x
										\right)
										\!
									\big\|_{1,+}
									\!-\!
									\frac{\delta}{2}
									\big\|
										\mT_{\frac{\gamma}{\delta}}^{+}
										\left(
											x
										\right)
										\!
									\big\|_2^2
									\!+\!
									\delta
									\cdot 
									x^T
									\mT_{\frac{\gamma}{\delta}}^{+}
									\left(
										x
									\right)
									\nn\\
							&=
									-\gamma
									\!\cdot\!
									\big\|
										\mT_{\frac{\gamma}{\delta}}^{+}
										\left(
											x
										\right)
										\!
									\big\|_{1}
									\!-\!
									\frac{\delta}{2}
									\big\|
										\mT_{\frac{\gamma}{\delta}}^{+}
										\left(
											x
										\right)
										\!
									\big\|_2^2
									\!+\!
									\delta\!
									\cdot\! 
									x^T
									\mT_{\frac{\gamma}{\delta}}^{+}
									(
										x
									)
	\end{align}
where the last step uses the fact that the output of $\mT_{\gamma}^{+}(\cdot)$ is always nonnegative so that $\|\mT_{\frac{\gamma}{\delta}}^{+}	\left(x\right)\|_{1,+} = \|\mT_{\frac{\gamma}{\delta}}^{+} \left(x\right)\|_{1}$.


\bibliographystyle{IEEEbib}
\bibliography{DistOpt,DictLearn}

\end{document}